\documentclass[journal, 10pt]{IEEETran} 
\usepackage{times}
\usepackage{amsmath,amssymb,float,arydshln,color}
\usepackage{psfrag,setspace,wrapfig,subfigure,float}
\usepackage[utf8]{inputenc}
\usepackage{pifont}
\usepackage{dsfont}
\usepackage{epsfig}
\usepackage{epstopdf}
\usepackage{graphicx}
\usepackage{scrextend}
\usepackage{hyperref}
\usepackage[ruled]{algorithm2e}
\usepackage{amsfonts}
\usepackage{cite}
\usepackage{hyperref}
\usepackage{url}
\usepackage{booktabs}       
\usepackage{hhline}

\usepackage[table]{xcolor}
\usepackage{multirow}
\usepackage{tabu}
\allowdisplaybreaks

\newtheorem{lemma}{{Lemma}}
\newtheorem{assumption}{{ Assumption}}
\newtheorem{theorem}{{Theorem}}

\def\tran{^{\mathsf{T}}}

\DeclareMathOperator*{\argmin}{arg\,min}
\newcommand{\bp}{ \begin{proof}}
	\newcommand{\ep}{\end{proof} }

\newcommand{\Ex}{\mathbb{E}\hspace{0.05cm}}

\newcommand{\bm}[1]{\mbox{\boldmath $#1$}}

\newcommand{\be}{\begin{equation}}
\newcommand{\ee}{\end{equation}}
\newcommand{\bqq}{\begin{eqnarray}}
\newcommand{\eqq}{\end{eqnarray}}
\newcommand{\bal}{\begin{align}}
\newcommand{\eal}{\end{align}}
\newcommand{\bqn}{\begin{eqnarray*}}
	\newcommand{\eqn}{\end{eqnarray*}}
\newcommand{\nn}{\nonumber}
\newcommand{\ba}{\left[ \begin{array}}
	\newcommand{\ea}{\\ \end{array} \right]}
\newcommand{\qd}{\hfill{$\blacksquare$}}
\newcommand{\define}{\;\stackrel{\Delta}{=}\;}

\def\bsigma  {{\boldsymbol \sigma}}
\def\bphi  {{\boldsymbol \phi}}

\def\Pr {{\mathbb{P}}}

\def\tphi   {\widetilde{\boldsymbol \phi}}

\def\g{{\boldsymbol{g}}}

\def\n{{\boldsymbol{n}}}

\def\u{{\boldsymbol{u}}}

\def\w{{\boldsymbol{w}}}

\newcommand{\cF}{{\mathcal{F}}}

\newcommand{\cU}{{\mathcal{U}}}

\newcommand{\tw}{\widetilde{\boldsymbol{w}}}

\newcommand{\grad}{{\nabla}}

\def\bE{\mathbb{E}}
\def\filt{\boldsymbol{\mathcal{F}}}

\newcommand{\eq}[1]{\begin{align}#1\end{align}}
\newcommand{\beqn}{\begin{eqnarray}}
\newcommand{\eeqn}{\end{eqnarray}}

\newcommand{\nnb}{\nonumber \\}

\def\real{{\mathbb{R}}}

\def\doubleunderline#1{\underline{\underline{#1}}}

\def\Zint{{\mathchoice{\setbox1=\hbox{\sf Z}\copy1\kern-.75\wd1\box1}
		{\setbox1=\hbox{\sf Z}\copy1\kern-.75\wd1\box1}
		{\setbox1=\hbox{\scriptsize\sf Z}\copy1\kern-.75\wd1\box1}
		{\setbox1=\hbox{\scriptsize\sf Z}\copy1\kern-.75\wd1\box1}}}

\makeatletter
\def\hlinewd#1{%
	\noalign{\ifnum0=`}\fi\hrule \@height #1 \futurelet
	\reserved@a\@xhline}
\makeatother

\begin{document}
	\def\helvetica{phvr7t.tfm}
	\def\helveticaoblique{phvro7t.tfm}
	\def\helveticabold{phvb7t.tfm}
	\def\helveticaboldoblique{phvbo7t.tfm}
	
	\font\sfb=\helveticabold
	=\helveticaboldoblique
	\title{
		Variance-Reduced Stochastic
		Learning under Random Reshuffling
	}
	\author{Bicheng Ying, Kun Yuan, and Ali H. Sayed\vspace{-0.8cm}\thanks{B. Ying and K. Yuan are with the Department of Electrical and Computer Engineering, UCLA, Los Angeles. A. H. Sayed is with the School of Engineering, Swiss Federal Institute of Technology (EPFL), Switzerland. This work was supported in part by NSF grant CCF-1524250. Emails: \{ybc, kunyuan\}@ucla.edu, and ali.sayed@epfl.ch}}
	\maketitle
	\begin{abstract}\vspace{-0mm}
		Several useful variance-reduced stochastic gradient algorithms, such as SVRG, SAGA, Finito, and SAG, have been proposed to minimize empirical risks with linear convergence properties to the exact minimizer. The existing convergence results assume uniform data sampling with replacement. However, it has been observed in related works that random reshuffling can deliver superior performance over uniform sampling and, yet, no formal proofs or guarantees of exact convergence exist for variance-reduced algorithms under random reshuffling. This paper makes two contributions. First, it resolves this open  issue and provides the first theoretical guarantee of linear convergence under random reshuffling for SAGA; the argument is also adaptable to other variance-reduced algorithms. Second, under random reshuffling, the paper proposes a new amortized variance-reduced gradient (AVRG) algorithm with constant storage requirements compared to SAGA and with balanced gradient computations compared to SVRG. AVRG is also shown analytically to converge linearly.
	\end{abstract}\vspace{-1.5mm}
	\vspace{-0.5mm}
	
	\setlength{\abovedisplayskip}{1.2mm}
	\setlength{\belowdisplayskip}{1.2mm}
	\section{Introduction and Motivation}\vspace{-1mm}
	In recent years, several useful variance-reduced stochastic gradient algorithms have been proposed, including SVRG \cite{johnson2013accelerating}, SAGA \cite{defazio2014saga}, Finito \cite{defazio2014finito}, SDCA\cite{shalev2013stochastic}, and SAG\cite{roux2012stochastic}, with the intent of reaching the exact minimizer of an empirical risk. Under constant step-sizes and strong-convexity assumptions on the loss functions, these methods have been shown to attain linear convergence towards the exact minimizer when the data samples are uniformly sampled {\em with} replacement.
	

	However, it has been observed in related works \cite{ bottou2009curiously,recht2012toward,gurbuzbalaban2015random} that implementations that rely instead on random reshuffling (RR) of the data (i.e., sampling {\em without} replacement) achieve better performance than implementations that rely on uniform sampling with replacement. Under random reshuffling, the algorithm is run multiple times over the finite data set. Each run is indexed by the integer $t\geq 1$ and is referred to as an epoch. For each epoch, the original data is  reshuffled so that
	the sample of index $i$ becomes the sample of index ${\bsigma}^t(i)$, where the
	symbol ${\bsigma}$ is used to refer to  a uniform random permutation of the indices. 
	
	It was shown in \cite{gurbuzbalaban2015random} that random reshuffling under {\em decaying} step-sizes can accelerate the convergence rate of stochastic-gradient learning from $O(1/i)$ to $O(1/i^2)$ \cite{nesterov2013introductory,polyak1987introduction}, where $i$ is the iteration index. It was also shown in \cite{ying2017rr} that random reshuffling under small {\em constant} step-sizes, $\mu$, can boost the steady-state performance of these algorithms from $O(\mu)$-suboptimal to $O(\mu^2)$-suboptimal around a small neighborhood of the exact minimizer \cite{sayed2014adaptation}. A similar improvement in convergence rate and performance has been observed for the variance-reduced Finito algorithm \cite{defazio2014finito}. However, no formal proofs or guarantees of exact convergence exist for the class of variance-reduced algorithms {\em under random reshuffling}, i.e., it is still not known whether these types of algorithms are still guaranteed to converge to the exact minimizer  when RR is employed and under what conditions on the data. {\color{black} For example, in \cite{de2016efficient}, another variance-reduction algorithm is proposed under reshuffling; however, no proof of convergence is provided.} The closest attempts at proof are the useful arguments given in \cite{gurbuzbalaban2015convergence,Shamir2016Without}, which deal with special problem formulations. The work \cite{gurbuzbalaban2015convergence} deals with the case of incremental aggregated gradients, which corresponds to a deterministic version of RR for SAG, while the work \cite{Shamir2016Without} deals with SVRG in the context of ridge regression problems using regret analysis. \vspace{-0.3mm}
	
	Motivated by these considerations, this paper makes two contributions. First, it resolves the open convergence issue and provides the first theoretical proof and guarantee of linear convergence to the exact minimizer under random reshuffling for  SAGA. While the argument is easily adaptable to a wider class of variance-reduced implementations, we illustrate the technique in this work for the  SAGA algorithm due to space limitations. A second contribution is that, under random reshuffling, we will propose a new amortized variance-reduced gradient (AVRG) algorithm with two  benefits: it has constant storage requirements in comparison to SAGA, and it has balanced gradient computations in comparison to SVRG. The balancing in computations is attained by amortizing the full gradient calculation across all iterations.  AVRG  is also shown analytically to converge linearly. \vspace{-0.3mm}
	
	
	In preparation for the analysis, we review briefly some of the conditions and notation that are relevant. We consider a generic empirical risk function $J(w):\real^{M}\rightarrow \real$, which is defined as a sample average of loss values over a possibly large but finite training set of size $N$:
	\eq{\label{prob-emp-into}
		w^\star \define \argmin_{w\in \real^M}\;  J(w) \define \frac{1}{N}\sum_{n=1}^{N}Q(w;x_n),
	}
	where the $\{x_n\}_{n=1}^N$ represent training data samples.\vspace{-0mm}
	
	\begin{assumption}[\sc Loss function]
		\label{assumption.1}
		The loss function $Q(w;x_n)$ is convex, differentiable, and has a $\delta$-Lipschitz continuous gradient, i.e., for every $n=1,\ldots,N$ and any $w_1, w_2 \in \real^M$: \vspace{-0.3mm}
		\eq{\label{eq-ass-cost-lc-e}
			\|\grad_w Q(w_1;x_n) - \grad_w Q(w_2;x_n) \| \le \delta \|w_1-w_2\|
		}
		where $\delta > 0$. We also assume that the empirical risk $J(w)$ is $\nu$-strongly convex, namely,\vspace{-1mm}
		\eq{\label{eq-ass-cost-sc-e}
			\hspace{-1mm}\Big(\grad_w J(w_1)- \grad_w J(w_2)\Big)\tran (w_1 - w_2) &\ge \nu \|w_1-w_2\|^2
		}\vspace{-4mm}
	\end{assumption}

	\section{SAGA with random reshuffling }\vspace{-0mm}
	We consider the SAGA algorithm \cite{defazio2014saga} in this work, while noting that our analysis can be easily extended to other versions of variance-reduced algorithms; for example, we shall illustrate how the approach applies to the new variant designated by the acronym  AVRG.  We list the SAGA algorithm without the proximal step but incorporate random reshuffling into the description of the algorithm. We explain the symbols and the operation of the algorithm following the table. In the listing below, note that, random quantities are being denoted  in boldface font,  which will be our standard convention in this work.\\
	\noindent\rule{0.48\textwidth}{1.4pt}
	{{\bf SAGA with Random Reshuffling \cite{defazio2014saga}}}\vspace*{-2mm}\\
	\rule{0.48\textwidth}{1pt}
	{\bf \small Initialization}:
	\(
	\w_0^0 = 0, \nabla Q(\bphi_{0,n}^0 ;x_n) = 0,\, n=1,2,\ldots,N.
	\)\\
	{\bf\small Repeat $t=0,1,2\ldots, T$} (epoch):\\
	\mbox{\quad\;\;generate a random permutation function\;}$\bsigma^t(\cdot).$\\
	\mbox{\bf \small \quad\;\; Repeat $i=0,1,\ldots N-1$} (iteration):\vspace{-2mm}
	\begin{align}
	\n =&\, \bsigma^t(i+1)\\[-1mm]
	\w_{i+1}^t =&\, \w_{i}^t - \mu \Big[\nabla Q(\w_{i}^t;x_{\n}) - \nabla Q(\bphi^t_{i,\n}{};x_{\n}) \nn\\[-1mm]
	&\hspace{1.2cm} {} + \frac{1}{N}\sum_{n=1}^N\nabla Q(\bphi^t_{i,n};x_{n})\Big]\label{main.step}\\[-1mm]
	\bphi^t_{i+1,\n} =&\, \w_{i+1}^t,\;\;{\rm and}\;\;\bphi^t_{i+1,n} =\, \bphi^t_{i,n},\;\;{\rm for}\;n\neq\n
	\label{observe.1}\end{align}
	\mbox{\bf \small \quad\;\; End}
	\eq{
		\w_0^{t+1} =&\, \w_N^{t},\;\;\bphi^{t+1}_0=\bphi^{t}_N\hspace{3.5cm}
	}
	{\bf\small End} \vspace{-1.5mm}\\
	\rule{0.48\textwidth}{1pt}\vspace{-2mm}

	\subsection{Operation of the Algorithm}
\begin{figure}[htb]
		\centering
		\includegraphics[scale=0.32]{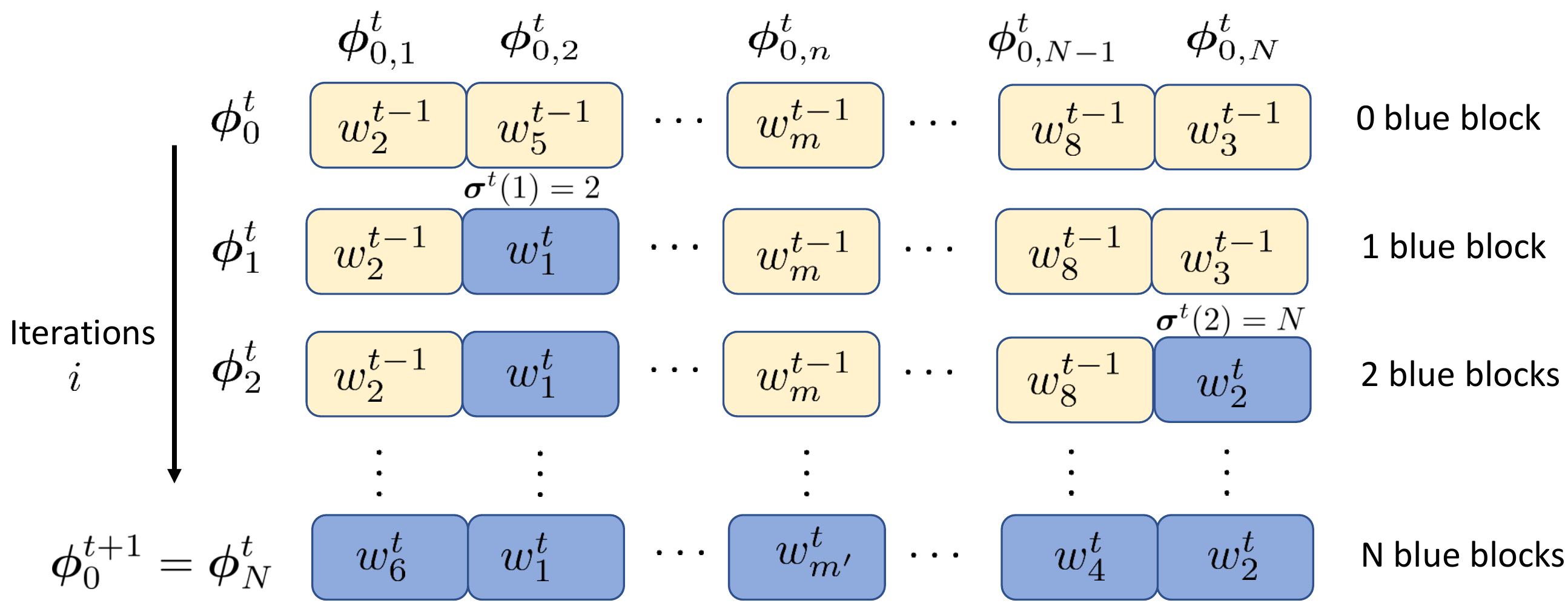}\vspace{-2mm}
		\caption{\small{An illustration of the evolution of the history variables $\{\bphi^t_{i,n}\}$.}}\vspace{-2.5mm}
		\label{fig.auxiliary.variable}
	\end{figure}

	Note that the algorithm runs a total of $T$ times over the data of size $N$. For each  run $t$, the original data $\{x_n\}_{n=1}^N$ is first randomly reshuffled so that the sample of index $i+1$ becomes the sample of index $\n={\bsigma}^{t}(i+1)$ in that run To facilitate the understanding of the algorithm, we associate a block matrix $\bm{\Phi}^t$ with each run, as illustrated in Fig.~\ref{fig.auxiliary.variable}.  This matrix is only introduced for visualization purposes. We denote the block rows of $\bm{\Phi}^t$ by  $\{\bm{\phi}_i^t\}$; one for each iteration $i$. Each block row $\bm{\phi}_i^t$ has size $M\times N$, with its entries generated by the SAGA recursion:
	\be
	\bm{\phi}_i^{t}\define \ba{ccccccc}\bm{\phi}_{i,1}^t &\vline& \bm{\phi}_{i,2}^t&\vline &\ldots& \vline & \bm{\phi}_{i,N}^t\ea\;\;\;(\mbox{\rm $i-$th block row})
	\ee
	We can therefore view $\bm{\Phi}^t$ as consisting of cells $\{\bm{\phi}^t_{i,n}\}$, each having  the same $M\times 1$ size as the minimizer $w^{\star}$. At every iteration $i$, one random cell in the $(i+1)-$th block row is populated by the iterate $\w^t_{i+1}$; the column location of this random cell is determined by the value of $\n$.\vspace{-0mm}
	
	We refer to Fig. 1 and explain in greater detail how the cells in the figure are updated. These cells play the role of history variables. To begin with, at iteration $i=0$, the cells in the first block row $\bm{\phi}_0^t$ will contain a randomly reshuffled version of all iterates $\{\w_1^{t-1}, \w_{2}^{t-1},\ldots,\w_{N}^{t-1}\}$ generated during the previous run of index $t-1$. A random sample of index $\n=\bm{\sigma}^t(1)$ is selected. Assume this value turns out to be $\n=2$. Then, as indicated in the blue cell in the second block row in the figure, the second cell of
	$\bm{\phi}_1^t$ is updated to $\w_{1}^t$ while all other cells in this row remain invariant. Moving to iteration $i=1$, a new random sample of index $\n=\bm{\sigma}^t(2)$ is selected. Assume this value turns out to be $\n=N$. Then, as indicated again in the third block row in the figure, the last cell of $\bm{\phi}_2^t$ is updated to $\w_2^t$ while all other cells in this row remain invariant. The process continues in this manner, by populating the cell corresponding to location $\n$ in the $i-$th block row. By the end of iteration $N$, all cells of $\bm{\phi}_N^t$ would have been populated by the iterates $\{\w_i^t\}$ generated during the $t-$th run. Observe that, since uniform sampling {\em with} replacement is used, then all weight iterates $\{\w_i^t\}$, from $i=1$ to $i=N$ will appear in $\bm{\phi}_N^t$. These iterates appear randomly shuffled in the last row in the figure and they constitute the initial value for $\bm{\phi}_0^{t+1}$ for the next run.\vspace{-1mm}
	
	\subsection{Properties of the History Variables}\vspace{-1mm}
	Several useful observations can be drawn from Fig. \ref{fig.auxiliary.variable}. These properties will be useful in the convergence proof in subsequent sections.\vspace{-0mm}
	
	{\bf Observation 1}: At the start of each epoch $t$, the components $\{\bphi^t_{0,n}\}_{n=1}^N$ correspond to a permutation of the weight iterates from the previous run, $\{\w^{t-1}_i\}_{i=1}^N$. $\hfill\Box$
	
	{\bf Observation 2}: At the beginning of the $i-$th iteration of an epoch $t$, all components of indices
	$\{\bm{\sigma}^t(m)\}_{m=1}^i$ will be set to weight iterates obtained during the $t-$th run, namely, $\{\w_m^t\}_{m=1}^i$, while the remaining $N-i$ history positions will have values from the previous run, namely, $\{\w^{t-1}_{k_n}\}_{n=1}^{N-i}$ for some values $k_n\in \{1,2,\ldots,N\}$.$\hfill\Box$

	{\bf Observation 3}: At the beginning of the $i-$th iteration of an epoch $t$, it holds that
	\eq{
		\bphi^{t}_{i,\n} = \bphi^{t}_{0,\n}  ,\;\;\;\;\;{\rm where} \; \n\in\bsigma^t(i+1{:}N)
	}
	where $\bsigma^t(i+1{:}N)$ represents the selected indices for future iterations $i+1$ to $N$. This property holds because, under random reshuffling, sampling is performed without replacement.$\hfill\Box$
	
	Using these observations, the following two results can be established.\vspace{-0mm}
	\begin{lemma}[\sc Distribution of history variables]
		\label{lemma.1}
		Conditioned on the previous $t-1$ epochs, each history variable $\bphi_{i,\n}^t$ has the following probability distribution at the beginning of the $i-$th ($i<N$) iteration of epoch $t$:\vspace{-1mm}
		\eq{\label{uniform-dist}
			\Pr(\bphi_{i,\n}^t|{\bm \cF}_0^{t}) =
			\left\{
			\begin{aligned}
				1/N, \;\;\;&\bphi_{i,\n}^t=\w^{t-1}_1	\\[-0.5mm]
				1/N, \;\;\;&\bphi_{i,\n}^t=\w^{t-1}_2	\\[-2.2mm]
				\vdots\;\;&\\[-1mm]
				1/N,\;\;&\bphi_{i,\n}^t=\w^{t-1}_N
			\end{aligned}
			\right.,\;\;\;\;{\rm for} \; \n\in\bsigma^t(i+1{:}N)
		}
		where ${\bm \cF}_0^{t}$ is the collection of all information before iteration 0 at epoch $t$.\vspace{-0mm}
	\end{lemma}
	{\bf Proof}: See Appendix \ref{app.lemma.1}. 
	\qd \vspace{-.5mm}
	
	\begin{lemma}[\sc Second-order moment of $\bphi^t_{i,n}$]\label{lemma.2}
		The aggregate second-order moment of each history variable $\bphi_{i,n}^t$ is equal to:
		\eq{
			\Ex\left[\sum_{n=1}^N \|\bphi_{i,n}^t\|^2\right] =  \sum_{n'=1}^i \Ex\|\w_{n'}^t\|^2+\frac{N-i}{N}\sum_{n=1}^N \Ex\|\w^{t-1}_{n}\|^2 \label{290jg32.ge}
		}\vspace{-1mm}
	\end{lemma}\vspace{-0.5mm}
	{\bf Proof}: See Appendix \ref{app.lemma.2}. 
	\qd \vspace{-0.5mm}
	
	For comparison purposes, the results obtained so far do not hold for implementations that involve sampling the data {\em with} replacement. For example, in that case (\ref{290jg32.ge}) would be replaced instead by the following expression derived in  \cite{defazio2014saga}:
	\eq{
		\Ex\left[\sum_{n=1}^N \|\bphi_{i,n}^t\|^2\right] =  \Ex\|\w^t_i\|^2 + \frac{N-1}{N} \sum_{n=1}^N\Ex\|\bphi_{i-1,n}^t\|^2\label{leads.to.2}
	}
	This result is similar to \eqref{290jg32.ge} only for $i=1$. However, observe that (\ref{leads.to.2}) involves variables $\{\bphi_{i-1,n}^t\}$ on the right-hand side, instead of the variables $\{\w^{t-1}_{n}\}$ that appear in (\ref{290jg32.ge}). This is because random reshuffling updates every history variable during each run, while uniform sampling may leave some variables $\bphi_{i-1,n}^t$ untouched. As we are going to illustrate in later experiments, this difference helps explain why SAGA under random reshuffling  tends to have faster convergence rate.
	\vspace{-2mm}
	
	\subsection{Biased Nature of the Gradient Estimator}\vspace{-0mm}
	Before launching into the convergence analysis of the variance-reduced algorithm, we first highlight one useful observation, namely, that  it is not necessary to insist on {\it unbiased} gradient estimators for proper operation of stochastic-gradient algorithms. To see this, let us examine first the SAGA implementation assuming uniform data sampling {\em with} replacement. In a manner similar to (\ref{main.step}), the SAGA algorithm in this case
	will employ the following  modified gradient direction: \vspace{-1mm}
	\eq{
		\widehat{g}_{\u}(\w_{i}^t)\define& \nabla Q(\w_{i}^t;x_{\u}) - \nabla Q(\bphi^t_{i,\u};x_{\u}) \nn\\
		&\;\;+ \frac{1}{N}\sum_{n=1}^N\nabla Q(\bphi^t_{i,n};x_{n})
	}
	where the subscript $\u$ is used to denote a uniformly distributed random variable, $\u\sim \cU[1,N]$.
	As a result, this modified gradient  satisfies the {\em unbiasedness} property \cite{defazio2014saga}:
	\eq{
		\Ex_{\u}[\widehat{g}_{\u}(\w_{i}^t) |{\bm \cF}^t_{i}] =\nabla J(\w_i^t)
	}
	where ${\bm \cF}^t_{i}$ denotes the collection of all available information before iteration $i$ at epoch $t$. However, this property no longer holds under random reshuffling! This is because data is now sampled {\em without} replacement and the selection of one index becomes dependent on the selections made prior to it. Specifically, let 
	\eq{
	\widehat{\g}_{\n}(\w_i^t) 
	\define& \nabla Q(\w_{i}^t;x_{\n}) - \nabla Q(\bphi^t_{i,\n};x_{\n}) \nn\\
	&\;\;+ \frac{1}{N}\sum_{n=1}^N\nabla Q(\bphi^t_{i,n};x_{n}) \label{rr.saga.grad}
	}
	denote the stochastic gradient that is employed by the SAGA recursion (\ref{main.step}). It then holds that
	\setlength{\belowdisplayskip}{1.2mm}
	\eq{
		&\hspace{-3mm}\Ex_{\n}\big[\widehat{g}_{\n}(\w_{i}^t) |{\bm \cF}^t_{i}\big]\hspace{-1mm}\nn\\
		&\;\;=\hspace{-0.5mm}
		\frac{1}{N-i}\hspace{-1mm}\sum_{n\notin\bsigma^t(1{:}i)}\hspace{-2mm}\left(\nabla Q(\w_{i}^t;x_{n}) - \nabla Q(\bphi^t_{i,n};x_{n})\right)\hspace{-0.5mm} \nn\\[-1mm]
		&\;\;\;\;\;\;+ \frac{1}{N}\sum_{n=1}^N\nabla Q(\bphi^t_{i,n};x_{n})\label{wgweio.ge}
	}
	\noindent where $\n \!=\! \bsigma^t(i\!+\!1)$ {\color{black} and we exploit the uniform property of random reshuffling when expanded the expectation\cite{ying2017rr}
	\eq{
			\Pr[\bsigma^t(i+1)=n \,|\, \bsigma^t(1\colon i)] =&\,\left\{
		\begin{aligned}
			\frac{1}{N-i},\;\; & n \notin  \bsigma^t(1{:}i)\\
			0\;\;\;,\;\; & n\in   \bsigma^t(1{:} i)
		\end{aligned}
		\right.
	}
	where $\bsigma^t(1{:} i)$ represents the collection of permuted indices for the original samples numbered $1$ through $i$.
	}. It is not hard to see that the expression on the right-hand side of (\ref{wgweio.ge}) is generally~different from $\nabla J(\w_i^t)$. Consequently, the gradient estimate that is employed by SAGA under random reshuffling in (\ref{main.step}) is not an unbiased estimator for the true gradient. Nevertheless, we will establish two useful facts in the following sections. First, the gradient estimate (\ref{rr.saga.grad}) becomes asymptotically unbiased when the algorithm converges, as $t\rightarrow\infty$. Second, the biased gradient estimation does not harm the convergence rate because we will observe later that SAGA under random reshuffling  actually converges faster than SAGA under uniform sampling with replacement {\color{black} in the simulations.}\vspace{-2.mm}

	\subsection{Convergence Analysis}\vspace{-.5mm}
	The analysis employs two supporting lemmas. To begin with, we relate the starting iterates for two successive epochs as follows by summing all gradient terms in (\ref{main.step}) over $i$:
	\eq{
		\w_{0}^{t+1}=\;&\w_N^t\nn\\
		 =\;&\w_0^t-\mu \sum_{i=0}^{N-1} \left[\nabla Q(\w_{i}^t;x_{\n^t_i})-\nabla Q(\bphi_{i,\n_i^t}^t;x_{\n^t_i})\right.\nn\\
		&\left.{}\hspace{20mm} +\frac{1}{N}\sum_{n=1}^N\nabla Q(\bphi_{i,n}^t;x_{n}) \right]
	}
	{\color{black} where we are using the notation  $\n^t_i=\bsigma^t(i+1)$.} As already alluded to, one main difficulty in the analysis is the fact that the gradient estimate is biased. For this reason, we shall compare against the gradient at the start of the epoch:
	\eq{
		\w_{0}^{t+1}
		\stackrel{(a)}{=}&\,\w_0^t-\mu N \underline{\nabla J(\w_{0}^t)}\nn\\
		&\;\;\;+\mu\sum_{i=0}^{N-1} \Big[\nabla Q(\bphi_{0,\n^t_i}^t;x_{\n^t_i}) -\frac{1}{N}\sum_{n=1}^N \doubleunderline{\nabla Q(\bphi_{0,n}^t;x_{n})} \Big]\nn\\[-0.5mm]
		&\;\;\;-\mu\sum_{i=0}^{N-1} \Bigg[\nabla Q(\w_{i}^t;x_{\n^t_i}) - \underline{\nabla Q(\w_{0}^t;x_{\n^t_i})} \nn\\
		&\;\;\;\;\;\;\;+\frac{1}{N}\sum_{n=1}^N\Big(\nabla Q(\bphi_{i,n}^t;x_{n})-\doubleunderline{\nabla Q(\bphi_{0,n}^t;x_{n})}\Big)\Bigg]\nn\\[-2.mm]
		\stackrel{(b)}{=}&\,\w_0^t-\mu N \nabla J(\w_{0}^t)\nn \\[-1mm]
		&\;\;\;\,-\mu\sum_{i=0}^{N-1} \Big[\nabla Q(\w_{i}^t;x_{\n^t_i}) - \nabla Q(\w_{0}^t;x_{\n^t_i})\nn\\
		&\;\hspace{9mm}+\frac{1}{N}\sum_{n=1}^N\Big(\nabla Q(\bphi_{i,n}^t;x_{n})-\nabla Q(\bphi_{0,n}^t;x_{n})\Big)\Big]\nn\\[-3mm]\label{38jg3.f3}
	}
	where in step (a) we added and subtracted $\{\nabla Q(\w_{0}^t;x_{n})\}$ and $\{\nabla Q(\bphi_{0,n}^t;x_{n})\}$, and we also changed the notation $\nabla Q(\bphi_{i,\n_i^t}^t;x_{\n_i^t})$ into $\nabla Q(\bphi_{0,\n_i^t}^t;x_{\n_i^t})$ because of observation 3; in step (b) we exploited the random reshuffling property that each index is selected only once, i.e.,
	{\color{black}
	\eq{
		\sum_{i=0}^{N-1} \nabla Q(\bphi_{0,\n^t_i}^t;x_{\n^t_i}) 
		\equiv \sum_{n=0}^{N-1} \nabla Q(\bphi_{0,n}^t;x_{n}) 
	}}
	
	We also need to appeal to a second recursion (within epoch $t$). By moving $\w_i^t$ in (\ref{main.step}) to the left-hand side and computing the squared norm,
	we obtain:\vspace{-0mm}
	\eq{
		&\|\w^t_{i+1} - \w^t_{i}\|^2\nn\\
		&=\mu^2\Big\|\nabla Q(\w_{i}^t;x_{\n}) - \nabla Q(\bphi_{i,\n}^t;x_{\n}) + \frac{1}{N}\sum_{n=1}^N\nabla Q(\bphi_{i,n}^t;x_{n})\Big\|^2\nn\\[-1mm]
		&\stackrel{(a)}{\leq}     3\mu^2\Big\|\nabla Q(\w_{i}^t;x_{\n})-\nabla Q(\w^t_0;x_{\n})\Big\|^2 \nn\\
		&\hspace{8mm} + 3\mu^2\Big\| \nabla Q(\bphi_{i,\n}^t;x_{\n})-\nabla Q(\w^{t-1}_N;x_{\n})\Big\|^2\nn\\[-0.5mm]
		&\hspace{8mm} + 3\mu^2\Big\|\frac{1}{N}\sum_{n=1}^N[\nabla Q(\bphi_{i,n}^{t};x_{n})-\nabla Q(w^\star;x_{n})]\Big\|^2\nn\\
		&\stackrel{(b)}{\leq}  3\delta^2\mu^2 \|\w_{i}^t-\w_0^t \|^2 \hspace{-0.5mm}+\hspace{-0.5mm} 3\delta^2\mu^2 \|\w_N^{t-1}- \bphi_{i,\n}^t\|^2\hspace{-0.5mm}\nn\\
		&\;\hspace{6mm}+\hspace{-0.5mm} \frac{3\delta^2\mu^2}{N}\sum_{n=1}^N\|\bphi_{i,n}^t - w^\star \|^2
		\label{eq.inner.diff}\vspace{-2mm}
	}
	where in step (a) we first added and subtracted $\nabla Q(\w^t_0;x_\n)$ and used the fact that $\frac{1}{N}\sum_{n=1}^N\nabla Q(w^\star;x_{n})=0$; then, we employed Jensen's inequality; and step (b) is because of the assumed Lipschitz condition (\ref{eq-ass-cost-lc-e}). Using (\ref{38jg3.f3}) and (\ref{eq.inner.diff}), and further introducing the error quantity
	$\tw_i^t=w^\star - \w_i^t$, we can establish the following auxiliary lemmas.\vspace{-0mm}
	\begin{lemma}[\sc Mean-square error recursion] \label{lemma.eror.quantity}
		The mean-square-error at the start of each epoch satisfies the following inequality recursion for step sizes $\mu\leq 1/(N\nu)$:
		\\[-4mm]
		\eq{
			&\hspace{-0mm}\Ex\|\tw_0^{t+1}\|^2 \label{eq.eror.quantity}\\
			&\;\leq \left(1-\frac{\mu\nu N - \mu^2N^2\delta^2}{1-\mu N\nu}\right) \Ex\|\tw_{0}^t\|^2\nn\\
			&\;\;\;{} + 4\mu\frac{\delta^2}{\nu}\left(\sum_{i=1}^{N-1}\Ex\|\w_{i}^t-\w_0^t\|^2 {}+ \sum_{n'=1}^{N-1}\Ex
			\|\w^{t-1}_N - \w^{t-1}_{n'}\|^2\right)\nn\vspace{-0mm}
		}
	\end{lemma}\vspace{-0.1mm}
	{\bf Proof}: See Appendix \ref{app.lemma.error.quantity}.
	\qd\\[2mm]
	Roughly, the above result shows that the mean-square error across epochs evolves according to a dynamics that is determined by the scaling factor 
	\be
	\alpha \define \left.1-\frac{\mu\nu N - \mu^2N^2\delta^2}{1-\mu N\nu}\right.
	\ee
	which is smaller than one for small $\mu$.
	In addition, there are two driving terms in (\ref{eq.eror.quantity}). We will refer
	$\sum_{i=1}^{N-1}\Ex\|\w_{i}^t-\w_0^t\|^2$ as the forward inner difference term and to  $\sum_{n'=1}^{N-1}\Ex
	\|\w^{t-1}_N - \w^{t-1}_{n'}\|^2$ as the backward inner difference term.
	\vspace{-0mm}
	
	\begin{lemma}[\sc Inner differences]\label{lemma.forward.backward}
		The forward inner difference satisfies:
		\eq{
			&\hspace{-0mm}\sum_{i=1}^{N-1}\hspace{-0.6mm}\Ex\|\w_i^t- \w_0^t\|^2\nn\\[-3mm]
			&\;\leq 5\delta^2\mu^2N^2\hspace{-0.8mm} \left(\hspace{-0.3mm}\sum_{i=1}^{N-1}\Ex\|\w_{i}^t-\w_0^t \|^2 \hspace{-0.6mm}+\hspace{-0.6mm} \sum_{i=1}^{N-1}\Ex\|\w_N^{t-1}- \w_{i}^{t-1}\|^2\hspace{-0.6mm}\right) \hspace{-0.9mm}\nn\\
			&\;\;\;\;\;\;\;{}+\hspace{-0.6mm} 3\delta^2\mu^2N^3\Ex\|\tw_0^t\|^2\label{eq.forward}
		}
		while the backward inner difference satisfies:
		\eq{
			&\hspace{-0mm}\sum_{i=1}^{N-1}\Ex\| \w_N^{t-1}-\w_i^{t-1}\|^2\nn\\[-3mm]
			&\;\leq\hspace{-0.3mm} 5\delta^2\mu^2N^2\!\hspace{-0.5mm} \left(\hspace{-0.3mm}\sum_{i=1}^{N-1}\!\Ex\!\|\w_{i}^{t-1}\hspace{-0.5mm}-\hspace{-0.5mm}\w_0^{t-1} \|^2\hspace{-0.3mm}\!+\!\!\hspace{-0.3mm} \sum_{i=1}^{N-1}\!\Ex\!\|\w_N^{t-2}- \w_{i}^{t-2}\|^2\hspace{-0.3mm}\!\right) \hspace{-0.8mm}\nn\\
			&\;\;\;\;\;\;{}+\hspace{-0.6mm} 3\delta^2\mu^2N^3\Ex\|\tw_0^{t-1}\|^2\label{eq.backward}
		}
	\end{lemma}\vspace{-0mm}
	{\bf Proof}: See Appendix  \ref{app.forward.backward}.
	\qd\\[2mm]
	Combining the above lemmas, we arrive at the following theorem.
	Let $\widetilde{\w}_0^t\define w^{\star} - \w_0^t$ and introduce the energy function:
	\eq{
		&V_{t+1}\define \Ex\|\tw_0^{t+1}\|^2 + \\ 
		&\;\;\;\;\;\;\;\frac{11}{16}\gamma\hspace{-0.6mm}\left(\hspace{-0.6mm}\frac{1}{N}\sum_{i=1}^{N-1}\Ex\|\w_i^{t+1}\!-\! \w_0^{t+1}\|^2 \!+\!\frac{1}{N}\sum_{i=1}^{N-1}\Ex\| \w_N^t\!-\!\w_i^t\|^2  \hspace{-0.6mm}\right)\nn
	}
	where $\gamma=9\mu\delta N$.\vspace{-0mm}
	
	\begin{theorem} [\sc Linear convergence of SAGA]\label{theorem.1}  For sufficiently small step-sizes, namely, for $\mu\leq\frac{\nu}{11\delta^2N}$, the quantity $V_{t+1}$ converges linearly:
		\eq{
			V_{t+1}\leq \alpha V_{t}
		}
		where 
		\eq{
			\alpha \hspace{-0.4mm}=\hspace{-0.4mm}\frac{1-\mu\nu N / 4}{1-27\delta^4 \mu^3 N^3/\nu}\hspace{-0.4mm} < \hspace{-0.4mm}1
		}
		It follows that $\Ex\|\tw_0^{t}\|^2\hspace{-0.3mm}\leq\hspace{-0.3mm}\alpha^t V_0$. 
	\end{theorem}
	{\bf Proof}: See Appendix \ref{app.theorem.1}.
	\qd\\[2mm]
	{\color{black} {\bf Remark}: To achieve an $\epsilon$-optimal solution, the number of iterations required is close to $O(\delta^2 / \nu^2 ) \log(1/\epsilon)$, which is slower than the rate proved under sampling without replacement in \cite{defazio2014saga}. The main reason is that the dependency between the samples makes it difficult to obtain a tight bound. As we will observe in the simulations later, in practice, the convergence can be faster than the original SAGA.}\vspace{-0mm}

	\section{Amortized Variance-Reduced Gradient (AVRG) Learning}\label{sec.avrg}\vspace{-1mm}
	One inconvenience of the SAGA implementation is its high storage requirement, which refers to the need to track the history variables $\{\bm{\phi}^t_{i,n}\}$ or the gradients for use in (\ref{main.step}). There is a need to store $O(N)$ variables. In big data applications, the size of $N$ can be prohibitive. The same storage requirement applies to the variant with reshuffling proposed in \cite{de2016efficient}. An alternative method  is the stochastic variance-reduced gradient (SVRG) algorithm \cite{johnson2013accelerating}, which is listed below (again with random reshuffling) for ease of reference.\\
	\begin{table}
	\noindent\rule{0.48\textwidth}{1.4pt}
	{{\bf SVRG with Random Reshuffling}\cite{johnson2013accelerating}}\vspace*{-2mm}\\
	\rule{0.48\textwidth}{1pt}
	{\bf \small Initialization}:
	\(
	\w_0^0 = 0.
	\)\\
	{\bf\small Repeat $t=0,1,2\ldots, T$} (epochs):\\
	\mbox{\quad\;\;$\nabla J(\w^t_0)\;=\;\displaystyle\frac{1}{N}\sum_{n=1}^N\nabla Q(\w_0^t;x_{n})$}\\
	\mbox{\quad\;\;generate a random permutation function\;}$\bsigma^t(\cdot).$\\
	\mbox{\bf \small \quad\;\; Repeat $i=0,1,\ldots N-1$} (iteration):
	\begin{align}
	\n =& \bsigma^t(i+1)\hspace{4.cm}\\
	\w_{i+1}^t =& \w_{i}^t \!-\! \mu \left[\nabla Q(\w_{i}^t;x_{\n}) \!-\! \nabla Q(\w_0^t;x_{\n}) \!+\! \nabla J(\w^t_0)\right]\label{main.svrg.step}
	\end{align}
	\mbox{\bf \small \quad\;\; End}
	\eq{
		\w_0^{t+1} =&\, \w_N^{t} \hspace{5.2cm}
	}
	{\bf\small End} \vspace{-1.5mm}\\
	\rule{0.48\textwidth}{1pt}\vspace{-1.5mm}
	\end{table}\vspace{-0mm}
	This method replaces the history variables $\{\bm{\phi}_{i,\n}^t\}$ of SAGA by a fixed initial condition $\w_0^t$ for each epoch. This simplification greatly reduces the storage requirement. However, each epoch in SVRG is  preceded by an aggregation step to compute a gradient estimate, which is time-consuming for large data sets. It also causes the operation of SVRG to become {\em unbalanced}, with a larger time interval needed before each epoch, and shorter time intervals needed within the epoch. 
	Motivated by these two important considerations, we propose a new {\em amortized} implementation, referred to as AVRG. This new algorithm removes the initial aggregation step from SVRG and replaces it by an estimate $\g^{t+1}$. This estimate is computed iteratively within the inner loop by re-using the gradient, $\nabla Q(\w^{t}_i;x_{\n})$, to reduce complexity.\vspace{-2.5mm}
	
	\begin{table}
		\noindent\rule{0.48\textwidth}{1.4pt}
		{{\bf AVRG with Random Reshuffling}}\vspace*{-2mm}\\
		\rule{0.48\textwidth}{1pt}
		{\bf \small Initialization}:
		\(
		\w_0^0 = 0,\;\;\g^0=0,\;\;\nabla Q(\w_0^0;x_n)\leftarrow 0,\, n=1,2,\ldots,N
		\)\\
		{\bf\small Repeat $t=0,1,2\ldots, T$} (epoch):\\
		\mbox{\quad\;\;generate a random permutation function\;}$\bsigma^t(\cdot)$, \mbox{\;\;\;\;\;\,set $\g^{t+1}=0$}\\
		\mbox{\bf \small \quad\;\; Repeat $i=0,1,\ldots N-1$} (iteration):
		\setlength{\abovedisplayskip}{.5mm}
		\setlength{\belowdisplayskip}{.5mm}
		\begin{align}
		\n =&\, \bsigma^t(i+1)\hspace{2.4cm}\\[-0.2mm]
		\;\;\w_{i+1}^t =&\, \w_{i}^t - \mu \left[\nabla Q(\w_{i}^t;x_{\n}) - \nabla Q(\w_0^t;x_{\n}) + \g^t\right]\label{eq.ns.svrg2}\\[-0.2mm]
		\;\;\g^{t+1}\leftarrow&\,\g^{t+1}+\frac{1}{N}\nabla Q(\w^{t}_i;x_{\n})\label{eq.ns.svrg3}
		\end{align}
		\mbox{\bf \small \quad\;\; End}\vspace{-0.1mm}
		\eq{
			\w_0^{t+1} =&\, \w_N^{t} \hspace{5.2cm}\vspace{-6mm}
		}
		{\bf\small End} \vspace{-1.7mm}\\
		\vspace{-2mm}\rule{0.48\textwidth}{1pt}\vspace{-5mm}
	\end{table}
	\begin{table*}\label{table.compare}
	\centering
	\caption{\normalsize Comparison of the variance-reduced implementations: SAGA, SVRG, SAG, and AVRG.}\vspace{-1.5mm}
	\begin{tabular}{|l||c|c|c|c|c|c|c|}
		\hline
		& SVRG\cellcolor[gray]{0.8} &\hspace{-1.5mm} SVRG+RR\hspace{-1.5mm}\cellcolor[gray]{0.8} & AVRG\cellcolor[gray]{0.8} & SAG\cellcolor[gray]{0.8}& SAGA \cellcolor[gray]{0.8} & \hspace{-1.5mm}SAGA+RR\hspace{-1.5mm} \cellcolor[gray]{0.8}\\
		\hhline{=======}
		{\small gradient computation per epoch}&\hspace{-1mm}$2.5N$\hspace{-1mm} &$2.5N$ &\cellcolor[gray]{0.92}$2N$&$N$&$N$&$N$\\
		\hline
		{\small extra storage requirement}& $O(1)$ & $O(1)$ &\cellcolor[gray]{0.92} $O(1)$ & $O(N)$ & $O(N)$&$O(N)$\\
		\hline
		{\small balanced gradient computation}&No &No &\cellcolor[gray]{0.92}Yes &Yes &Yes &Yes\\
		\hline
		{\small unbiased gradient estimator} &Yes &No &\cellcolor[gray]{0.92}No &No &Yes &No\\
		\hline
	\end{tabular}\vspace{-4.5mm}
\end{table*}
	
	\subsection{Useful Properties}\vspace{-1mm}
	Several properties stand out when we compare the proposed AVRG implementation with the previous algorithms. First, observe that the storage requirement for AVRG in each epoch is just the variables $\g^{t}$, $\g^{t+1}$, and $\w^t_0$, which is similar to SVRG and considerably less than SAGA. 
	
	Second, since the gradient vector $Q(\w^{t}_i;x_{\n})$ used in  (\ref{eq.ns.svrg3}) has already been computed in \eqref{eq.ns.svrg2},  every iteration $i$ will only require two gradients to be evaluated. Thus, the effective computation of gradients per epoch is smaller in AVRG than in SVRG.\vspace{-0.mm}
	
	Third, observe from  Eq. (\ref{eq.ns.svrg3}) how the estimated $\g^t$ is computed by averaging the loss values at successive iterates. This construction is feasible because of the use of random reshuffling. Under random reshuffling, the collection of gradients   $\{Q(\w^{t}_i;x_{\n})\}$ that are used  in (\ref{eq.ns.svrg3}) during each epoch will end up covering the entire set of data, $\{x_n\}_{n=1}^N$. This is not necessarily the case for operation under uniform sampling with replacement. Therefore, the AVRG procedure assumes the use of random reshuffling. We will simply refer to it as AVRG, rather than AVRG under RR.\vspace{-0.2mm}
	
	Fourth, unlike the SVRG algorithm, which requires a step to compute the full gradient, the AVRG implementation is amenable to decentralized implementations {\color{black}  (i.e., to fully-decentralized implementations with no master nodes).} and also to asynchronous operation \cite{NIPS2015_5821}. The unbalanced gradient computation in SVRG poses difficulties for fully-decentralized solutions \cite{konevcny2016federated, sayed2014adaptation,yuan2017efficient} (instead of master-slave model) and introduces idle times when multiple devices/agents with different amounts of data cooperate to solve an optimization problem. 
	The amenability to effective decentralized solutions is a powerful convenience of the AVRG framework and one main motivation for introducing it, as explained in the related work \cite{yuan2017efficient}.
	
	Finally, the modified gradient direction that is employed  in \eqref{eq.ns.svrg2} by AVRG has distinctive properties in relation to the modified gradient direction \eqref{main.step} in SAGA. To see this, we note that the gradient direction in \eqref{eq.ns.svrg2} can be written as\vspace{-1mm}
	\eq{
		\widehat{g}_\n(\w^t_i) \define& \nabla Q(\w_{i}^t;x_{\n}) - \nabla Q(\w^{t}_0;x_{\n}) \nn\\
		&\;\;\;+ \frac{1}{N}\sum_{n=0}^{N-1} \nabla Q(\w^{t-1}_n ; x_{\bsigma^{t-1}(n+1)}) \label{23mgo.d}
	}
	It is clear that even when the index $\n$ is chosen uniformly, the above vector cannot be an unbiased estimator for true gradient in general. What is more critical for convergence is that the modified gradient direction of an algorithm should satisfy the useful property that as the weight iterate gets closer to the optimal value, i.e., as $\|w^{\star}-\w^t_i\|\leq\epsilon$, for arbitrary small $\epsilon$ and large enough $t$, the modified and true gradients  will also get arbitrarily close to each. This property holds for (\ref{23mgo.d}) since\vspace{-0.5mm}
	\eq{
		&\hspace{-0mm}\|\widehat{g}_\n(\w^t_i) - \nabla J(w^{\star})\|\nn\\[-0mm]
		&\leq \left\|\nabla Q(\w_{i}^t;x_{\n}) - \nabla Q(\w^{t}_0;x_{\n})\right\| \nn\\
		&\;\;\;{}+ \left\|\frac{1}{N}\sum_{n=1}^{N} \nabla Q(\w^{t-1}_{n-1} ; x_{\bsigma^{t-1}(n)}) - \frac{1}{N}\sum_{n=1}^N \nabla Q(w^{\star} ; x_n)\right\|\nn\\[-1mm]
		&\leq \delta\|\w_{i}^t-w^{\star}\| + \delta\|\w^{t}_0-w^{\star}\| +\frac{\delta}{N}\sum_{n=1}^{N-1} \left\| \w^{t-1}_{n-1} -w^{\star}\right\|\nn\\
		&\leq3\delta \epsilon
	}
	where in the second inequality we exploited Jensen's inequality, the triangle inequality, Lipschitz assumption, and the fact that $\bsigma^{t-1}(n)$ corresponds to sampling without replacement. Because $\epsilon$ can be chosen arbitrary small, then $\widehat{g}_\n(\w^t_i)$ must approach the true gradient at $w^{\star}$.  This result implies the aforementioned asymptotic unbiasedness property of the gradient estimate. Actually, this property holds for all previous modified gradients in SAGA, SVRG, SAG, Finito, and AVRG. The work \cite{NIPS2015_5711} also discusses a case where there is an extra error term in the gradient calculation, {\color{black} which supports the observation that a small gradient bias does not necessarily harm convergence.} For ease of reference, Table~\ref{table.compare} compares the trade-offs between storage and computational complexity of different variance-reduced algorithms with and without random reshuffling.	
	
	\subsection{Convergence Analysis}\vspace{-1mm}
	The same approach used to establish the convergence of SAGA under RR is also suitable for AVRG. For this reason, we can be brief. First, similar to \eqref{38jg3.f3}, we derive the main recursion for one epoch:\vspace{-1mm}
	\eq{
		\w^{t+1}_0
		\hspace{-0.8mm}=&\w^t_0 - \mu  N \nabla J(\w_0^t)\label{eq.avrg.main}\\
		&\;\;\;{}-\mu \sum_{i=0}^{N-1} \big[\nabla Q(\w_{i}^t;x_{\n^t_i})- \nabla Q(\w^{t}_0;x_{\n^t_i})\big]\nn\\
		&\;\;\;{}+\mu \sum_{i=0}^{N-1}\Big[\nabla Q(\w^{t}_0;x_{\n_{i}^{t-1}})-\nabla Q(\w^{t-1}_i;x_{\n_i^{t-1}})\Big] \nn
	}
	where, for compactness of notation,  we introduce $\n^{t-1}_i= \bsigma^{t-1}(i+1)$. Second, similar to \eqref{eq.inner.diff}, we derive an inner difference recursion:
	\eq{
		&\hspace{-3mm}\| \w_{i+1}^t - \w_{i}^t\|^2\nn\\
		\;\;=&\; \mu^2\left\|\nabla Q(\w_{i}^t;x_{\n}) - \nabla Q(\w^{t}_0;x_{\n}) + \g^t \right\|^2\nn\\
		\leq&\; 3\mu^2\delta^2\!\left(\!\|\w^t_i-\w^t_0\|^2 \!+\! \frac{1}{N}\sum_{i=0}^{N-1}\|\w^{t-1}_i \!-\!\w^{t-1}_N\|^2+\|\tw^{t}_0\|^2\!\right) \label{2389hg.23g}
	}
	Next, we establish recursions related to $\tw^t_0$, and the forward and backward difference terms.\vspace{-0mm}
	
	\begin{lemma}[\sc Recursions for AVRG analysis] \label{lemma.avrg}
		The mean-square-error at the start of each epoch satisfies the following inequality for step-sizes  $\mu\leq 1/(N\nu)$:\vspace{-1mm}
		\eq{
			&\Ex\|\tw^{t+1}_0\|^2\label{38jgg3g}\\
			&\leq\left(1-\frac{\mu\nu N - \mu^2N^2\delta^2}{1-\mu N\nu}\right)\Ex\|\tw^{t}_0\|^2 \nn\\
			&\;\;\;\;\;\;{}+\! \frac{2\mu\delta^2 }{\nu} \!\left( \sum_{i=0}^{N-1}\Ex\| \w_{i}^t-\w^{t}_0\|^2 \!+\! \sum_{i=0}^{N-1}\Ex\| \w_{N}^{t-1}-\w^{t-1}_{i}\|^2\!\right)\nn
		}
		Moreover, the forward inner difference satisfies:\vspace{-0mm}
		\eq{
			&\hspace{-0mm}\sum_{i=0}^{N-1}\Ex\|\w_i^t - \w_0^t\|^2
			\leq\; 3\mu^2\delta^2N^2 \sum_{i=0}^{N-1}\Ex\|\w_{i}^t-\w_0^t \|^2\nn\\[-1mm]
			&\hspace{15mm}{} + \mu^2 \delta^2 N^2\left(\sum_{i=0}^{N-1}\Ex\|\w_N^{t-1}- \w_{i}^{t-1}\|^2 + N\Ex\|\tw_0^t\|^2\right)
		}
		while the backward inner difference satisfies:\vspace{-0mm}
		\eq{
			&\hspace{-0mm} \sum_{i=0}^{N-1}\Ex\| \w_{N}^{t}-\w^{t}_{i}\|^2 
			\leq 3\mu^2\delta^2N^2\sum_{i=0}^{N-1}\Ex\|\w^t_i-\w^t_0\|^2\\
			&\hspace{10mm}{}+ 3\mu^2\delta^2N^2 \left( \sum_{i=0}^{N-1}\Ex\|\w^{t-1}_i-\w^{t-1}_N\|^2 + N \Ex\|\tw_0^t\|^2 \right)\nn
		}
	\end{lemma}\vspace{-0.mm}
	{\bf Proof}: See Appendix \ref{app.lemma.avrg}.
	\qd
	\begin{figure*}[!h]
		\centering
		\includegraphics[scale=0.58]{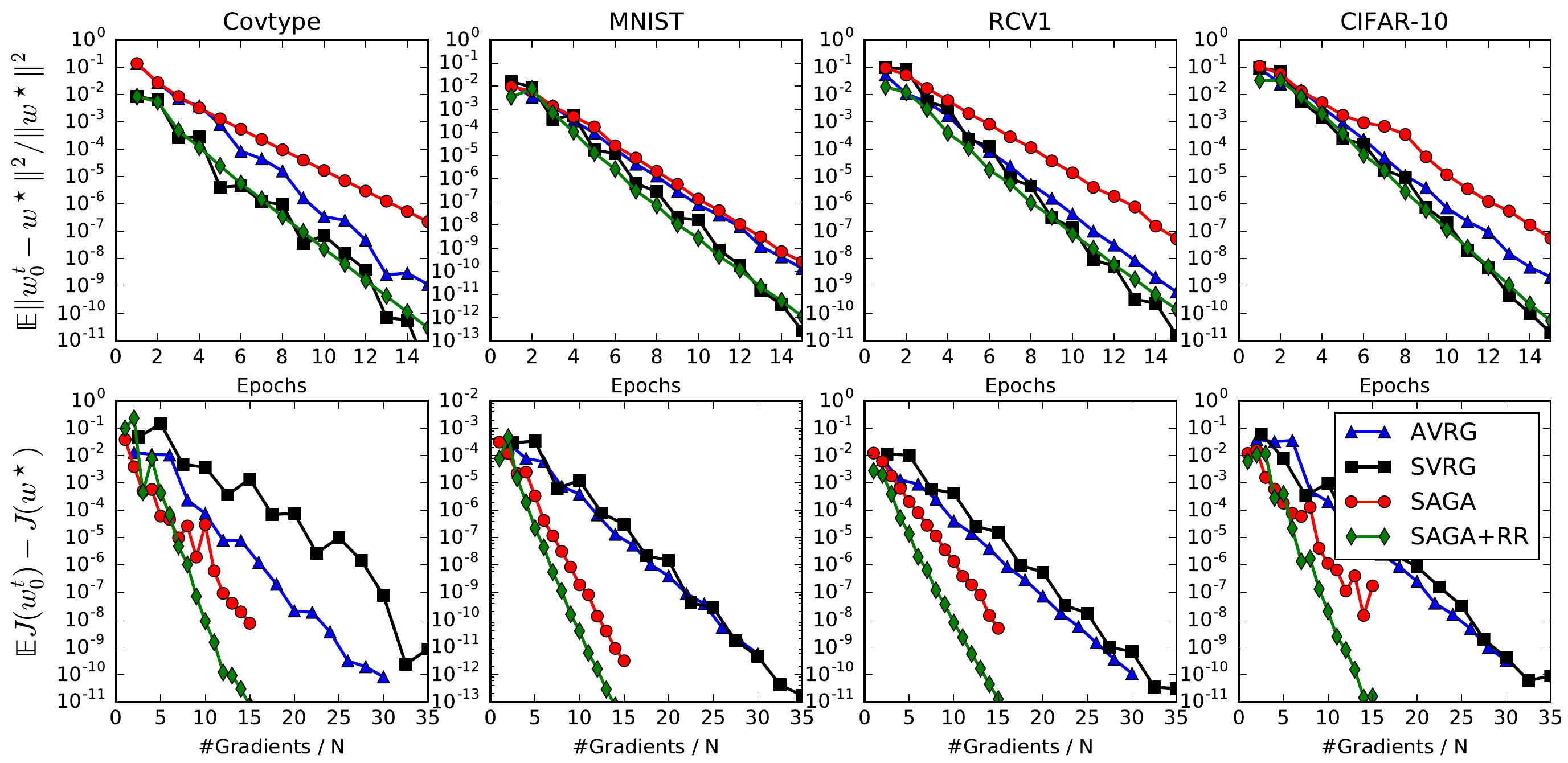}\vspace{-0.5mm}
		\caption{\small Comparison of various variance-reduced algorithms over three datasets: Covtype, MNIST, and CIFAR-10. The top three plots compare the relative mean-square-error performance versus the epoch index, $t$, while the bottom three plots compare the excess risk values versus the number of gradients computed. \vspace{-1mm}}
		\label{fig.exp_reuslt}
	\end{figure*}
	
	Likewise,  we introduce the energy function:
	\eq{
		V_{t+1}\define \Ex\|\tw_0^{t+1}\|^2 + \frac{13}{16}\gamma\left(\frac{1}{N}\sum_{i=1}^{N-1}\Ex\|\w_i^{t+1} - \w_0^{t+1}\|^2 \right.\nn\\
		\left.+\frac{1}{N}\sum_{i=1}^{N-1}\Ex\| \w_N^t-\w_i^t\|^2  \right)
	}
	where $\gamma=6\mu\delta N$, and state the relevant convergence theorem.
	\vspace{-0mm}
	
	\begin{theorem} [\sc Linear convergence of AVRG]\label{theorem.2}  For sufficiently small step-sizes, namely, for $\mu\leq\frac{\nu}{9\delta^2N}$, the quantity $V_{t+1}$ converges linearly:\vspace{-1mm}
		\eq{
			V_{t+1}\leq \alpha V_{t}
		}
		where 
		\be
			\alpha = \frac{1-\mu\nu N / 4}{1-18\delta^3 \mu^3 N^3/\nu} < 1
		\ee It follows that $\Ex\|\tw_0^{t}\|^2\leq \alpha^t V_0$.	
	\end{theorem}\vspace{-0mm}
	{\bf Proof}: See Appendix \ref{app.theorem.2}.
	\qd\\[2mm]
	{\color{black} {\bf Remark}: This is similar to the theorem for SAGA under RR except for the scaling coefficients. However, in practice, the convergence curve of AVRG will be different from the one of SAGA under RR. 
	\vspace{-0mm}
	
	\section{Simulation Results}\vspace{-0mm}
	In this section, we illustrate the convergence performance of various algorithms by numerical simulations. We consider the following regularized logistic regression problem:\vspace{-1mm}
	\eq{\label{xcn23bh}
		\min_w\quad J(w) =\;& \frac{1}{N}\sum_{n=1}^{N} Q(w;h_n,\gamma(n))\nn\\
		\define& \frac{1}{N}\sum_{n=1}^{N}\left(\frac{\rho}{2} \|w\|^2 + \ln\big(1+\exp(-\gamma(n) h_n\tran w)\big)\right)
	}
	where $h_n\!\in\!\real^M\!$ is the feature vector,  $\gamma(n)\!\!\in\!\! \{\pm1\}$ is the class label. In all our experiments, we set $\rho\! =\! 1/N$. The optimal $w^\star$ and the corresponding risk value are calculated by means of the Scikit-Learn package. We run simulations over four datasets: covtype.binary\footnote{\label{footnote.1}\url{http://www.csie.ntu.edu.tw/~cjlin/libsvmtools/datasets/}}, rcv1.binary\footref{footnote.1},
	MNIST\footnote{\url{http://yann.lecun.com/exdb/mnist/}}, and CIFAR-10\footnote{\url{http://www.cs.toronto.edu/~kriz/cifar.html}}. The last two datasets have been transformed into binary classification problems by considering data with labels 0 and 1, i.e., digital zero and one classes for MNIST and airplane and automobile classes for CIFAR-10. All features have been preprocessed and normalized to the unit vector \cite{xiao2014proximal}. The results are exhibited in Fig. \ref{fig.exp_reuslt}. To enable fair comparisons, we tune the step-size parameter of each algorithm for fastest convergence in each case. The plots are based on measuring the relative mean-square-error, $\Ex\|\w^t_0-w^\star\|^2 / \|w^\star\|^2,$ and the excess risk value, $\Ex J(\w^t_0) - J(w^\star).$ Two key facts to observe from these simulations are that 1) SAGA with RR is consistently faster than SAGA, and
	2) without the high memory cost of SAGA and without the unbalanced structure of SVRG, the proposed AVRG technique is able to match their performance reasonably well. Moreover, as we shall show in future work \cite{yuan2017efficient},  the AVRG technique enables effective distributed implementations.\vspace{-0mm}
	

	\section{Discussion and Future Work}\vspace{-0mm}
	The statements of Theorems \ref{theorem.1} and \ref{theorem.2} are similar. This suggests that the analysis approach is applicable to a wider class of variance-reduced implementations. The statements also suggest that these types of algorithms are able to deliver linear convergence for sufficiently small
	constant step-sizes. One useful extension for future study is to consider situations with non-smooth loss functions. It is also useful to note that the stability ranges and convergence rates derived from the theoretical analysis tend to be more conservative than what is actually observed in experiments. 
	\appendices
	
	\section{Proof of lemma \ref{lemma.1}}\label{app.lemma.1}
	For $\n=\bm{\sigma}^t(i+1)$ and any $\w^{t-1}_i$, $i=1,2,\ldots,N$, it holds that
	\eq{
		\Pr(\bphi_{i,\n}^t=\w^{t-1}_i|{\bm \cF}_0^{t} ) = &
		\sum_{\bsigma^t} \Pr(\bsigma^t) \Pr(\bphi_{i,\n}^t=\w^{t-1}_i|{\bm \cF}_0^{t}, \bsigma^t )
		\nn\\[-1mm]
		=&\sum_{\bsigma^t} \frac{1}{N!} \Pr(\bphi_{i,\n}^t=\w^{t-1}_i|{\bm \cF}_0^{t}, \bsigma^t )\nn\\
		=&  \sum_{\bsigma^t} \frac{1}{N!} \Pr(\bphi_{0,\n}^t=\w^{t-1}_i|{\bm \cF}_0^{t}, \bsigma^t )\nn\\
		=& \frac{1}{N!} \sum_{\bsigma^t} \mathbb{I}\,[\bphi_{0,\n}^t =\w^{t-1}_i |{\bm \cF}_0^{t}, \bsigma^t ]
	}
	The second equality is because all permutation sequences are equally probable; the third equality applies observation 3. The last equality follows from noting that, given $\filt^t_0$ and $\bsigma^t$,
	the quantity $\bphi_{0,\n}^t$~becomes a deterministic variable. In this case,
	the probability $\Pr(\bphi_{0,\n}^t|{\bm \cF}_0^{t}, \bsigma^t)$ is either 1 or 0. We therefore express it in terms of the indicator function, where the notation
	$\mathbb{I}[a]=1$ when the statement $a$ is true and is zero otherwise. Next note that there are $(N-1)!$ permutations $\bm{\sigma}^t$ with the $\n-$th position storing $w^{t-1}_i$. Substituting back, we get
	\eq{
		\Pr(\bphi_{i,\n}^t=\w^{t-1}_i|{\bm \cF}_0^{t} )	
		=& \frac{(N-1)!}{N!} \;=\;\frac{1}{N}}
	\qd
	\small
	\section{Proof of lemma \ref{lemma.2}} \label{app.lemma.2}
	\noindent Conditioning on the information in the past epochs:
	\eq{
		&\hspace{-1mm}\Ex\left[\left.\left(\sum_{n=1}^N \|\bphi_{i,n}^t\|^2 \right)\right| \filt^{t}_0 \right]\nn\\
		&\;= \Ex\left[\left.\left(\sum_{n\in \bsigma^t(1{:}i)} \|\bphi^t_{i,n}\|^2\right)\right| \filt^{t}_0 \right]+ \Ex\left[\left.\left(\sum_{n\notin \bsigma^t(1{:}i)} \|\bphi^t_{i,n}\|^2\right)\right | \filt^{t}_0 \right] \nn\\[-1mm]
		&\;= \Ex\left[\left.\left(\sum_{n'=1}^{i} \|\w_{n'}^t\|^2 \right)\right| \filt^{t}_0 \right] + \Ex\left[\left.\left(\sum_{i'=i+1}^N \|\bphi^t_{i,\bsigma^t(i')}\|^2\right)\right | \filt^{t}_0\right] \nnb
		&\;\overset{\eqref{uniform-dist}}{=} \sum_{n'=1}^{i}\bE\left[\|\w_{n'}^t\|^2 | \filt^{t}_0\right] + \sum_{i'=i+1}^{N}\frac{1}{N}\sum_{n=1}^{N}\|\w^{t-1}_{n}\|^2 \nn\\[-1mm]
		&\;= \sum_{n'=1}^{i}\bE\left[\|\w_{n'}^t\|^2 | \filt^{t}_0\right] + \frac{N-i}{N} \sum_{n=1}^{N}\|\w^{t-1}_{n}\|^2
	}
	Taking expectation over ${\bm \cF}^{t}_0$, we arrive at (\ref{290jg32.ge}).
	\qd
	
	\setlength{\abovedisplayskip}{.9mm}
	\setlength{\belowdisplayskip}{0.9mm}
	\section{Proof of lemma \ref{lemma.eror.quantity}}\label{app.lemma.error.quantity}\vspace{-1mm}
	By introducing the error quantity
	$\tw_i^t=w^\star - \w_i^t$, we easily arrive at the following recursion for the evolution of the error dynamics:
	\eq{
		\tw_0^{t+1}=&\;\tw_0^t+\mu N \nabla J(\w_{0}^t)\label{eq.main.recursion}\\
		&\;\;{}+\mu\sum_{i=0}^{N-1} \left[\nabla Q(\w_{i}^t;x_{\n_i^t}) - \nabla Q(\w_{0}^t;x_{\n_i^t})\right.\nn\\
		&\;\;\;\;\left.{}+\frac{1}{N}\sum_{n=1}^N\left(\nabla Q(\bphi_{i,n}^t;x_{n})-\nabla Q(\bphi_{0,n}^t;x_{n})\right)\right]\nn
	}
	Computing the conditional mean-square-error of both sides of \eqref{eq.main.recursion}, and appealing to Jensen's inequality, gives:
	\eq{
		&\hspace{-3mm}\Ex\left[\left\|\tw_{0}^{t+1}\right\|^2\,|\, {\bm \cF_{0}^t}\right]\nn\\
		\stackrel{(a)}{\leq}&\, \frac{1}{1-a}\left\|\tw_0^t+\mu N \nabla J(\w_{0}^t)\right\|^2\nn\\
		&+\frac{\mu^2}{a} \Ex\left\{\left\|\sum_{i=0}^{N-1} \left[\nabla Q(\w_{i}^t;x_{\n_{i}^t}) - \nabla Q(\w_{0}^t;x_{\n_{i}^t}) \right.\right.\right.\nn\\
		&\;\hspace{14mm}\left.\left.\left.+\frac{1}{N}\sum_{n=1}^N\left(\nabla Q(\bphi_{i,n}^t;x_{n})-\nabla Q(\bphi_{0,n}^t;x_{n})\right)\right]\right\|^2\!\Big| {\bm \cF_{0}^t}\right\}
		\nn\\
		\stackrel{(b)}{\leq}&\, \frac{1}{1-a}\left\|\tw_0^t+\mu N \nabla J(\w_{0}^t)\right\|^2\nn\\
		&+\frac{\mu^2N}{a} \sum_{i=0}^{N-1}\Ex\Bigg[\left\| \nabla Q(\w_{i}^t;x_{\n^t_i}) - \nabla Q(\w_{0}^t;x_{\n^t_i})\right.\nn\\
		&\;\hspace{14mm}\left.\left.+\frac{1}{N}\sum_{n=1}^N\left(\nabla Q(\bphi_{i,n}^t;x_{n})-\nabla Q(\bphi_{0,n}^t;x_{n})\right)\right\|^2
		\!\Big| {\bm \cF_{0}^t}\right]
		\nn\\
		\stackrel{(c)}{\leq}&\frac{1}{1-a}\left\|\tw_0^t+\mu N \nabla J(\w_{0}^t)\right\|^2\nn\\
		&\,\,+\frac{2\mu^2N}{a}\! \sum_{i=0}^{N-1}\!\Ex\!\left[\left\| \nabla Q(\w_{i}^t;x_{\n^t_i}) - \nabla Q(\w_{0}^t;x_{\n^{t}_i})\right\|^2\,|\, {\bm \cF_{0}^t}\right]\nn\\
		&\,\,+\frac{2\mu^2N}{a}\! \sum_{i=0}^{N-1}\!\Ex\!\left[\left\|\frac{1}{N}\sum_{n=1}^N\nabla Q(\bphi_{i,n}^t;x_{n})\!-\!\nabla Q(\bphi_{0,n}^t;x_{n})\right\|^2
		\!\Big| {\bm \cF_{0}^t}\right]\nn\\
		=&\,\frac{1}{1-a}\left\|\tw_0^t+\mu N \nabla J(\w_{0}^t)\right\|^2\nn\\
		&\,\,+\frac{2\mu^2N}{t}\! \sum_{i=1}^{N-1}\!\Ex\!\!\left[\left\| \nabla Q(\w_{i}^t;x_\n) - \nabla Q(\w_{0}^t;x_\n)\right\|^2\,|\, {\bm \cF_{0}^t}\right]\nn\\
		&\,\,+\frac{2\mu^2N}{a}\! \sum_{i=1}^{N-1}\!\Ex\!\!\left[\left\|\frac{1}{N}\sum_{n=1}^N\!\left(\nabla Q(\bphi_{i,n}^t;x_{n})\!-\!\nabla Q(\bphi_{0,n}^t;x_{n})\right)\right\|^2
		\!\Big| {\bm \cF_{0}^t}\right] \label{8t923.g8}
	}
	where step (a) follows from Jensen's inequality and $t$ can be chosen arbitrarily in the open interval $a\in (0,1)$; and steps (b) and (c) also follow from the following corollary of Jensen's inequality:
	\be
	\left\|\sum_{i=1}^N y_i\right\|^2=N^2\left\|\sum_{i=1}^N\frac{1}{N} y_i\right\|^2\leq N\sum_{i=1}^N\left\|y_i\right\|^2 \label{efj982f.g238}
	\ee
	We further know from the Lipschitz condition \eqref{eq-ass-cost-lc-e} that:
	\eq{
		\Ex\!\left[\left\|\nabla Q(\w_{i}^t;x_\n) - \nabla Q(\w_{0}^t;x_\n)\right\|^2\Big| {\bm \cF_{0}^t}\right]\leq \delta^2 \Ex\left[\|\w_{i}^t-\w_0^t\|^2\Big| {\bm \cF_{0}^t}\right]
		\label{f9j30.1}
	}
	and
	\eq{
		&\Ex\left[\left\|\frac{1}{N}\sum_{n=1}^N\left(\nabla Q(\bphi_{i,n}^t;x_{n})-\nabla Q(\bphi_{0,n}^t;x_{n})\right)\right\|^2
		\,\Big|\, {\bm \cF_{0}^t}\right]\nn\\
		&\stackrel{(a)}{=} \Ex\left[ \left\|\frac{1}{N}\sum_{n=1}^i\nabla Q(\w^{t}_{n};x_{\bsigma^t(n)})-\nabla Q(\bphi^{t}_{0,\bsigma^t(n)};x_{\bsigma^t(n)})\right\|^2\,\Big|\, {\bm \cF_{0}^t}\right]\nn\\
		&\stackrel{(b)}{\leq} \frac{i\delta^2}{N^2}\sum_{n=1}^i\Ex\left[ \|\w^{t}_{n}-\bphi^t_{0,\bsigma^t(n)}\|^2\,|\,{\bm \cF}^{t}_0\right]\nn\\
		&= \frac{i\delta^2}{N^2}\sum_{n=1}^i\Ex\left[ \|\w^{t}_{n}-\w_0^t + \w^{t-1}_N-\bphi^t_{0,\bsigma^t(n)}\|^2\,|\,{\bm \cF}^{t}_0\right]\nn\\
		&\stackrel{(c)}{\leq} \frac{i\delta^2}{N^2}\!\sum_{n=1}^i\!\Big(2\Ex\!\left[ \|\w^{t}_{n}-\w^{t}_{0}\|^2|{\bm \cF}^{t}_0\right] \!+\!
		2\Ex\!\left[ \|\w^{t-1}_{N}-\bphi^t_{0,\bsigma^t(n)}\|^2|{\bm \cF}^{t}_0\right] \Big)\nn\\
		&= \frac{i\delta^2}{N^2}\left(\sum_{n=1}^i 2\Ex\!\left[ \|\w^{t}_{n}-\w^{t}_{0}\|^2\,|\,{\bm \cF}^{t}_0\right] +
		\frac{2}{N}\sum_{n'=1}^N \|\w^{t-1}_{N}-\w^{t-1}_{n'}\|^2\right)\label{f9j30.2}
	}
	where step (a) holds because of observation 2, steps (b) and (c) apply Jensen's inequality; and the last equality is because of uniform random reshuffling. Next, using the strong-convexity of the empirical risk, we have that
	\eq{
		&\hspace{-4mm}\left\|\tw^t_0 + \mu N \nabla J(\w_0^t)\right\|^2\nn\\[1mm]
		=& \|\tw^t_0\| + \mu^2 N^2\| \nabla J(\w_0^t)\|^2+ 2 \mu N(\tw^t_0)\tran \nabla J(\w_0^t)\nn\\[1mm]
		\leq& \|\tw^t_0\| + \mu^2 N^2\delta^2 \|\tw_0^t\|^2\nn\\[1mm]
		&\hspace{5mm}{}- 2 \mu N(\w^t_0-w^\star)\tran (\nabla J(\w_0^t) - \nabla J(\w^\star) )\nn\\[1mm]
		\leq& (1-2\mu\nu N + \mu^2 N^2\delta^2 )\|\tw_0^t\|^2
		\label{f9j30.3}
	}
	Substituting (\ref{f9j30.1}), (\ref{f9j30.2}), and (\ref{f9j30.3}) into (\ref{8t923.g8}) and letting $a=\mu N\nu$, assuming $\mu\leq 1/(N\nu)$, we get
	\eq{
		&\hspace{-1mm}\Ex\left[\|\tw_{0}^{t+1}\|^2\,|\, {\bm \cF_{0}^t}\right]\nn\\
		&\;\leq\left(\frac{1-2\mu\nu N + \mu^2N^2\delta^2}{1-\mu N\nu}\right)\|\tw_{0}^t\|^2 \nn\\
		&\;\;\;\;\;\;+ 2\mu\frac{\delta^2}{\nu}\sum_{i=1}^{N-1}\Ex\left[\|\w_{i}^t-\w_0^t\|^2\,|\, {\bm\cF}_0^t\right]
		\nn\\
		&\;\hspace{5mm} {}+2\mu\frac{\delta^2}{\nu}\sum_{i=1}^{N-1}\frac{i}{N^2}\left(\sum_{n=1}^i2\Ex\left[\|\w_n^t-\w_0^t\|^2\,|\, {\bm\cF}_0^t\right] \right.\nn\\
		&\hspace{32mm}\left.{}+\frac{2}{N} \sum_{n'=1}^N
		\|\w^{t-1}_N - \w^{t-1}_{n'}\|^2\right)\nn\\
		&\;\stackrel{(a)}{=} \left(1-\frac{\mu\nu N - \mu^2N^2\delta^2}{1-\mu N\nu}\right)\|\tw_{0}^t\|^2\nn\\
		&\;\;\;\;\;\; + 2\mu\frac{\delta^2}{\nu}\sum_{i=1}^{N-1}\Ex\left[\|\w_{i}^t-\w_0^t\|^2\,|\, {\bm\cF}_0^t\right]
		\nn\\
		&\;\hspace{5mm} {}+2\mu\frac{\delta^2}{\nu}\sum_{n=1}^{N-1}\sum_{i=n}^{N-1}\frac{i}{N^2}\Bigg(2\Ex\left[\|\w_n^t-\w_0^t\|^2\,|\, {\bm\cF}_0^t\right] \nn\\
		&\hspace{38mm}\left.{}+\frac{2}{N} \sum_{n'=1}^N
		\|\w^{t-1}_N - \w^{t-1}_{n'}\|^2\right)\nn\\
		&\;\stackrel{(b)}{\leq} \left(1-\frac{\mu\nu N - \mu^2N^2\delta^2}{1-\mu N\nu}\right)\|\tw_{0}^t\|^2 \nn\\
		&\;\;\;\;\;\;\;+ 2\mu\frac{\delta^2}{\nu}\sum_{i=1}^{N-1}\Ex\left[\|\w_{i}^t-\w_0^t\|^2\,|\, {\bm\cF}_0^t\right]
		\nn\\
		&\;\hspace{6mm} {}+2\mu\frac{\delta^2}{\nu}\sum_{n=1}^{N-1}\frac{1}{2}\Bigg(2\Ex\left[\|\w_n^t-\w_0^t\|^2\,|\, {\bm\cF}_0^t\right]\nn\\
		&\hspace{30mm}\left.+\frac{2}{N} \sum_{n'=1}^N
		\|\w^{t-1}_N - \w^{t-1}_{n'}\|^2\right)\nn\\
		&\;\leq\left(1-\frac{\mu\nu N - \mu^2N^2\delta^2}{1-\mu N\nu}\right) \|\tw_{0}^t\|^2 \nn\\
		&\hspace{5mm}{}+ 4\mu\frac{\delta^2}{\nu}\sum_{i=1}^{N-1}\!\Ex\!\left[\|\w_i^t-\w_0^t\|^2| {\bm\cF}_0^t\right]\!+2\mu\frac{\delta^2}{\nu} \!\sum_{n'=1}^N\!
		\|\w^{t-1}_N - \w^{t-1}_{n'}\|^2\nn\\
		&\;\leq \left(1-\frac{\mu\nu N - \mu^2N^2\delta^2}{1-\mu N\nu}\right) \|\tw_{0}^t\|^2 \nn\\
		&\hspace{5mm}+ 4\mu\frac{\delta^2}{\nu}\!\left(\sum_{i=1}^{N-1}\!\Ex\!\left[\|\w_i^t-\w_0^t\|^2\,|\, {\bm\cF}_0^t\right] {}+ \sum_{n'=1}^{N-1}
		\|\w^{t-1}_N - \w^{t-1}_{n'}\|^2\right)\label{last.step.s}
	}
	where in step (a) and in several similar steps later, we are using the equality:
	\eq{
		\sum_{i=1}^{N-1} \sum_{n=1}^i f(n,i) \equiv \sum_{n=1}^{N-1} \sum_{i=n}^{N-1} f(n,i) \label{eq.exchange.sum}
	}
	As for step (b), the factor $\frac{1}{2}$ is because:
	\eq{
		\sum_{i=n}^{N-1} \frac{i}{N^2}= \frac{(N-n)(N-n-1)}{2N^2}\leq \frac{1}{2},\;\;\;\; 1\leq n\leq N-1
	}
	The last step (\ref{last.step.s}) is unnecessary; it is used to introduce symmetry into the expression and facilitate the treatment. Taking expectation over the past history ${\bm\cF}_0^t$ leads to \eqref{eq.eror.quantity}.
	
	\section{Proof of lemma \ref{lemma.forward.backward}} \label{app.forward.backward}
	Using \eqref{eq.inner.diff}, we can establish an upper bound for any inner difference based on $\w^t_0$ as follows:
	\eq{
		&\hspace{-3mm}\|\w_{i}^t-\w_{0}^t\|^2 \nn\\
		=&\, \|\w_i^t -\w_{i-1}^t + \w_{i-1}^t-\cdots - \w_0^t\|^2\nn\\
		=&\, i^2\left\|\frac{1}{i}\left(\w_i^t -\w_{i-1}^t + \w_{i-1}^t-\cdots - \w_0^t\right)\right\|^2\nn\\
		\leq &\, i \sum_{m=0}^{i-1} \|\w_{m+1}^t - \w_m^t\|^2\nn\\
		\stackrel{\eqref{eq.inner.diff}}{\leq}& 3\delta^2\mu^2 i\sum_{m=0}^{i-1}\!\left(\|\w_{m}^t-\w_0^t \|^2 \!+\! \|\w_N^{t-1}- \bphi_{m,\n}^t\|^2 \!+\! \frac{1}{N}\sum_{n=1}^N\|\tphi_{m,n}^t \|^2\!
		\right) \label{289jg3}
	}
	where $\tphi_{m,n}^t\! \define\! w^\star - \bphi_{m,n}^t$.
	It is important to remark here that now $\n=\bsigma^t(m+1)$, i.e., $\n$ is always associated with the index before it. Summing over $i$, we have
	\eq{
		&\hspace{-3mm}\sum_{i=1}^{N-1}\|\w_i^t - \w_0^t\|^2\nn\\
		&\leq 3\delta^2\mu^2 \sum_{i=1}^{N-1} i \sum_{m=0}^{i-1}\Bigg(\|\w_{m}^t-\w_0^t \|^2 + \|\w_N^{t-1}- \bphi_{m,\n}^t\|^2\nn\\
		&\hspace{28mm}\;\;\;\left. {}+ \frac{1}{N}\sum_{n=1}^N\|\tphi_{m,n}^t \|^2
		\right)\nn\\
		&\stackrel{(\ref{eq.exchange.sum})}{=} 3\delta^2\mu^2 \sum_{m=0}^{N-2}\sum_{i=m+1}^{N-1}i\Bigg(\|\w_{m}^t-\w_0^t \|^2 + \|\w_N^{t-1}- \bphi_{m,\n}^t\|^2\nn\\
		&\hspace{29mm}\;\;\;\left. {}+ \frac{1}{N}\sum_{n=1}^N\|\tphi_{m,n}^t \|^2
		\right)\nn\\
		&\stackrel{(a)}{\leq} \frac{3}{2}\delta^2\mu^2N^2 \sum_{m=0}^{N-2}\Bigg(\|\w_{m}^t-\w_0^t \|^2 + \|\w_N^{t-1}- \bphi_{m,\n}^t\|^2\nn\\
		&\hspace{23mm}\;\;\;\left. {}+\frac{1}{N}\sum_{n=1}^N\|\tphi_{m,n}^t \|^2\right)
		\nn\\
		&\stackrel{(b)}{=} \frac{3}{2}\delta^2\mu^2N^2 \sum_{m=0}^{N-2}\Bigg(\|\w_{m}^t-\w_0^t \|^2 + \|\w_N^{t-1}- \w_{m+1}^{t-1}\|^2\nn\\
		&\hspace{23mm}\;\;\;\left. {}+\frac{1}{N}\sum_{n=1}^N\|\tphi_{m,n}^t \|^2\right)
		\nn\\
		&= \frac{3}{2}\delta^2\mu^2N^2 \left(\sum_{m=0}^{N-2}\|\w_{m}^t-\w_0^t \|^2 + \sum_{i=1}^{N-1}\|\w_N^{t-1}- \w_{i}^{t-1}\|^2\right.\nn\\
		&\hspace{18mm}\;\;\;\left. {}+\sum_{m=0}^{N-2}\frac{1}{N}\sum_{n=1}^N\|\tphi_{m,n}^t \|^2\right)\nnb
		&\leq \frac{3}{2}\delta^2\mu^2N^2 \left(\sum_{i=1}^{N-1}\|\w_{i}^t-\w_0^t \|^2+ \sum_{i=1}^{N-1}\|\w_N^{t-1}- \w_{i}^{t-1}\|^2\right.\nn\\
		&\hspace{18mm}\;\;\;\left. {}+\sum_{i=0}^{N-2}\frac{1}{N}\sum_{n=1}^N\|\tphi_{i,n}^t \|^2\right)
		\label{23ns999}
	}
	where step (a) is because $\sum_{i=m+1}^{N-1}i$ is bounded by $\frac{N^2}{2}$, and step (b) uses the fact that $\bm{\phi}_{i,\n}^t=\w_{m+1}^t$ by construction. Then, computing the conditional expectation, we get:
	\eq{
		&\hspace{-5mm}\sum_{i=1}^{N}\Ex\left[\|\w_i^t - \w_0^t\|^2\,|\, {\bm \cF}^{t}_0\right]\vspace{-3mm}\nn\\\vspace{-3mm}
		\leq&\,\frac{3}{2}\delta^2\mu^2N^2 \left(\sum_{i=1}^{N-1}\Ex\left[\|\w_{i}^t-\w_0^t \|^2\,|\, {\bm \cF}^{t}_0\right] + \sum_{i=1}^{N-1}\|\w_N^{t-1}- \w_{i}^{t-1}\|^2\right.\nn\\
		&\hspace{18mm}\left.{}+ \sum_{i=0}^{N-2}\frac{1}{N}\sum_{n=1}^N\Ex\left[\|\tphi_{i,n}^t \|^2\,|\, {\bm \cF}^{t}_0\right]\right) \label{g89h23.g3}
	}
	To bound the last term, we first separate it into two quantities:
	\eq{\label{23bs9}
		\Ex\left[\|\tphi_{i,n}^t \|^2\,|\, {\bm \cF}^{t}_0\right]=\;& \Ex\left[\|\tphi_{i,n}^t - \tw_0^t + \tw_0^t\|^2\,|\, {\bm \cF}^{t}_0\right]\nn\\
		\leq\;& 2\Ex\left[\|\bphi_{i,n}^t - \w_0^t\|^2\,|\, {\bm \cF}^{t}_0\right]+2\|\tw_0^t\|^2
	}
	Using an argument similar to Lemma \ref{lemma.2}, we can establish that:
	\eq{
		&\hspace{-6mm}\Ex\left[\sum_{n=1}^N\|\bphi_{i,n}^t - \w_0^t\|^2\,\Big|\, {\bm \cF}^{t}_0\right]\nn\\
		=&\sum_{n=1}^i\Ex\left[\|\w_{i}^t - \w_0^t\|^2\,\Big|\, {\bm \cF}^{t}_0\right]+\frac{N-i}{N} \sum_{n'=1}^N\|\w_N^{t-1}-\w^{t-1}_{n'}\|^2\label{3989jg/3}
	}
	Combining results (\ref{23bs9}) and (\ref{3989jg/3}), we can bound the last term of \eqref{g89h23.g3}:
	\eq{
		&\sum_{i=0}^{N-2}\frac{1}{N}\sum_{n=1}^N\Ex\left[\|\tphi_{i,n}^t \|^2\,|\, {\bm \cF}^{t}_0\right]
		\nn\\
		&\leq\;
		\sum_{i=0}^{N-1}\frac{2}{N}\left(\sum_{n=1}^i\Ex\left[\|\w_{i}^t - \w_0^t\|^2\,|\, {\bm \cF}^{t}_0\right]\right.\nn\\
		&\hspace{15mm}\;\;\;\left. {}+\frac{N-i}{N} \sum_{n'=1}^N\|\w_N^{t-1}-\w^{t-1}_{n'}\|^2 \right)+2N\|\tw_0^t\|^2\nn\\
		&\leq\;\frac{2}{N}\sum_{i=0}^{N-1}\sum_{n=1}^i\Ex\left[\|\w_{i}^t - \w_0^t\|^2\,|\, {\bm \cF}^{t}_0\right]\nnb
		&\;\;\;\;\;\;+\frac{N+1}{N}\sum_{n'=1}^N\|\w_N^{t-1}-\w^{t-1}_{n'}\|^2+2N\|\tw_0^t\|^2\nn\\
		&\leq\;2\sum_{i=0}^{N-1}\Ex\left[\|\w_{i}^t - \w_0^t\|^2\,|\, {\bm \cF}^{t}_0\right]\nn\\
		&\;\;\;\;\;+2\sum_{n'=1}^N\|\w_N^{t-1}-\w^{t-1}_{n'}\|^2+2N\|\tw_0^t\|^2\nn\\
		&=\,2\!\sum_{i=1}^{N-1}\!\Ex[\|\w_{i}^t \!-\! \w_0^t\|^2| {\bm \cF}^{t}_0]+2\sum_{n'=1}^{N-1}\|\w_N^{t-1}\!-\!\w^{t-1}_{n'}\|^2
		\!+\!2N\|\tw_0^t\|^2
	}
	Substituting back into (\ref{g89h23.g3}), we have:
	\eq{
		&\hspace{-3mm}\sum_{i=1}^{N-1}\Ex\left[\|\w_i^t - \w_0^t\|^2\,|\, {\bm \cF}^{t}_0\right]\nn\\
		\leq&\;\frac{3}{2}\delta^2\mu^2N^2 \left(\sum_{i=1}^{N-1}\Ex\left[\|\w_{i}^t-\w_0^t \|^2\,|\, {\bm \cF}^{t}_0\right] + \sum_{i=1}^{N-1}\|\w_N^{t-1}- \w_{i}^{t-1}\|^2\right.\nn\\
		&\hspace{16mm}{}+2\sum_{i=1}^{N-1}\Ex\left[\|\w_{i}^t - \w_0^t\|^2\,|\, {\bm \cF}^{t}_0\right]\nn\\
		&\hspace{15mm}\left.{}+2\sum_{n'=1}^{N-1}\|\w_N^{t-1}-\w^{t-1}_{n'}\|^2+2N\|\tw_0^t\|^2\right)\nn\\
		\leq&\;\frac{9}{2}\delta^2\mu^2N^2 \left(\sum_{i=1}^{N-1}\!\Ex\!\left[\|\w_{i}^t-\w_0^t \|^2\,|\, {\bm \cF}^{t}_0\right] \!+\! \sum_{i=1}^{N-1}\|\w_N^{t-1}- \w_{i}^{t-1}\|^2\right)\nn\\
		&\;\;\;\; + 3\delta^2\mu^2N^3\|\tw_0^t\|^2\nn\\
		\leq&\;5\delta^2\mu^2N^2 \left(\sum_{i=1}^{N-1}\Ex\![\|\w_{i}^t-\w_0^t \|^2\,|\, {\bm \cF}^{t}_0] + \sum_{i=1}^{N-1}\!\|\w_N^{t-1}- \w_{i}^{t-1}\|^2\right) \nn\\
		&\;\;\;\;+ 3\delta^2\mu^2N^3\|\tw_0^t\|^2
	}
	Taking expectation over the filtration leads to (\ref{eq.forward}).
	
	Next, following similar arguments, we have the following for backward inner difference term:
	\eq{&\hspace{-5mm}\|\w_{N}^{t-1}-\w_{i}^{t-1}\|^2 \nn\\
		=&\; \|\w_N^{t-1} -\w_{N-1}^{t-1} + \w_{N-1}^{t-1}-\cdots - \w_i^{t-1}\|^2\nn\\
		\leq\; & (N-i) \sum_{m=i}^{N-1}\|\w_{m+1}^{t-1} - \w_{m}^{t-1}\|^2\nn\\[-1mm]
		\stackrel{(\ref{eq.inner.diff})}{\leq}&\;  3\delta^2\mu^2(N-i) \sum_{m=i}^{N-1}\Bigg(\|\w_{m}^{t-1}-\w_0^{t-1} \|^2 + \|\w_N^{t-2}- \bphi_{m,\n}^{t-1}\|^2\nn\\
		&\;\;\hspace{26mm}\left.{}+ \frac{1}{N}\sum_{n=1}^N\|\tphi_{m,n}^{t-1} \|^2
		\right)\label{13jgds.ge}
	}
	where $\tphi_{m,n}^{t-1}\! \define\! w^\star - \bphi_{m,n}^{t-1}$ and now $\n=\bsigma^{t-1}(m+1)$. Summing over $i$, we have
	\eq{\label{23ns999-3}
		&\sum_{i=1}^{N-1}\|\w_N^{t-1} - \w_i^{t-1}\|^2\nn\\
		&\leq 3\delta^2\mu^2 \sum_{i=1}^{N-1} (N-i) \sum_{m=i}^{N-1}\Bigg(\|\w_{m}^{t-1}-\w_0^{t-1} \|^2 + \|\w_N^{t-2}- \bphi_{m,\n}^{t-1}\|^2\nn\\
		&\hspace{37mm}\left.{}+ \frac{1}{N}\sum_{n=1}^N\|\tphi_{m,n}^{t-1} \|^2
		\right)\nn\\
		&= 3\delta^2\mu^2 \sum_{m=1}^{N-2}\sum_{i=m}^{N-1}(N-i)\Bigg(\|\w_{m}^{t-1}-\w_0^{t-1} \|^2 + \|\w_N^{t-2}- \bphi_{m,\n}^{t-1}\|^2\nn\\
		&\hspace{37mm}\left.{}+ \frac{1}{N}\sum_{n=1}^N\|\tphi_{m,n}^{t-1} \|^2
		\right)\nn\\
		&\leq \frac{3}{2}\delta^2\mu^2N^2 \sum_{m=1}^{N-2}\Bigg(\|\w_{m}^{t-1}-\w_0^{t-1} \|^2 + \|\w_N^{t-2}- \bphi_{m,\n}^{t-1}\|^2\nn\\
		&\hspace{26mm}\left.{}+ \frac{1}{N}\sum_{n=1}^N\|\tphi_{m,n}^{t-1} \|^2
		\right)\nn\\
		&= \frac{3}{2}\delta^2\mu^2N^2 \sum_{m=1}^{N-2}\Bigg(\|\w_{m}^{t-1}-\w_0^{t-1} \|^2 + \|\w_N^{t-2}- \w_{m+1}^{t-2}\|^2\nn\\
		&\hspace{26mm}\left.{}+ \frac{1}{N}\sum_{n=1}^N\|\tphi_{m,n}^{t-1} \|^2
		\right)
		\nn\\
		&= \frac{3}{2}\delta^2\mu^2N^2 \Bigg(\sum_{i=1}^{N-2}\|\w_{i}^{t-1}-\w_0^{t-1} \|^2 + \sum_{i=2}^{N-1}\|\w_N^{t-2}- \w_{i}^{t-2}\|^2\nn\\
		&\hspace{20mm}\left.{}+ \sum_{i=1}^{N-2}\frac{1}{N}\sum_{n=1}^N\|\tphi_{i,n}^{t-1} \|^2\right) \nnb
		&\le \frac{3}{2}\delta^2\mu^2N^2 \Bigg(\sum_{i=1}^{N-1}\|\w_{i}^{t-1}-\w_0^{t-1} \|^2 + \sum_{i=1}^{N-1}\|\w_N^{t-2}- \w_{i}^{t-2}\|^2\nn\\
		&\hspace{20mm}\left.{}+ \sum_{i=0}^{N-2}\frac{1}{N}\sum_{n=1}^N\|\tphi_{i,n}^{t-1} \|^2\right)
	}
	The above result is similar to (\ref{23ns999}) with $t$ replaced by $t-1$.  Therefore, the same procedure can now be followed to arrive at (\ref{eq.backward}).
	
	\section{Proof of theorem \ref{theorem.1}}\label{app.theorem.1}
	To simplify the notation, we introduce the symbols:
	\eq{
		a_t^2\!\define \!\frac{1}{N}\sum_{i=1}^{N-1}\Ex\|\w_i^t-\w_0^t \|^2, \;\;
		b_{t-1}^2\!\define\! \frac{1}{N}\sum_{i=1}^{N-1}\Ex\|\w_N^{t-1}- \w_{i}^{t-1}\|^2
	}
	Then, the results of the previous three lemmas can be rewritten in the form:
	\eq{
		\Ex\|\tw_0^{t+1}\|^2\!\leq&\! \left(\!1-\frac{\mu\nu N - \mu^2N^2\delta^2}{1-\mu N\nu}\right)\! \Ex\!\|\tw_0^t\|^2 \!+\! 4\mu N \frac{\delta^2}{\nu}(a_t^2+b_{t-1}^2) \label{z23n99}\\
		a_{t+1}^2\leq &\; 5\delta^2\mu^2N^2(a_{t+1}^2+b_{t}^2) + 3\delta^2 \mu^2 N^2\Ex\|\tw_0^{t+1}\|^2 \label{1389jg2}\\
		b_t^2\leq &\; 5\delta^2\mu^2N^2(a_t^2+b_{t-1}^2) + 3\delta^2 \mu^2 N^2\Ex\|\tw_0^t\|^2 \label{1389jg2.1}
	}
	
	We can simplify these relations further by recognizing certain bounds. To begin with, note that
	\eq{
		1-\frac{\mu\nu N - \mu^2N^2\delta^2}{1-\mu N\nu} &= 1 \!-\! \frac{\mu \nu N - \mu^2 N^2 \nu^2 + \mu^2 N^2 \nu^2 - \mu^2 N^2 \delta^2 }{1 - \mu N \nu} \nnb
		&= 1 \!-\! \mu \nu N + \frac{\mu^2 N^2 \delta^2 \,-\, \mu^2 N^2 \nu^2}{1-\mu N \nu} \nnb
		&= 1 \!-\! \frac{3\mu \nu N}{4} \!-\! \left( \frac{\mu \nu N}{4} \!-\! \frac{\mu^2 N^2 \delta^2 - \mu^2 N^2 \nu^2}{1-\mu N \nu} \right) \nnb
		&\le 1 \!-\! \frac{3\mu \nu N}{4}\label{23ns99}
	}
	where the last inequality holds when
	\eq{
		1-\mu N\nu >0,	\frac{\mu \nu N}{4} - \frac{\mu^2 N^2 \delta^2 - \mu^2 N^2 \nu^2}{1-\mu N \nu}\geq0 \nn\\
		 \Longleftrightarrow \ \mu \le \min\left\{\frac{1}{N\nu},\,\frac{\nu}{N(4\delta^2 - 3\nu^2)} \right\}\nn\\
		\label{3289hg2.d}
	}
	Since $\nu\leq \delta$, we can replace (\ref{3289hg2.d}) by the sufficient condition
	\eq{
		\boxed{{\rm Condition\ \#1:}\;\; \mu \leq \frac{\nu}{4\delta^2N}}
		\label{23ns3299}
	}
	Under this condition, and substituting \eqref{23ns99} into \eqref{z23n99}, we get
	\eq{
		\Ex\|\tw_0^{t+1}\|^2 \leq&\; \left(1 - \frac{3\mu \nu N}{4}\right) \Ex\|\tw_0^t\|^2 + 4\mu N \frac{\delta^2}{\nu}(a_t^2+b_{t-1}^2) \label{12389f.2}
	}
	Let $\gamma$ denote an arbitrary positive scalar that we are free to choose. Multiplying relations (\ref{1389jg2}) and (\ref{1389jg2.1}) by $\gamma$ and adding to \eqref{12389f.2} we obtain:
	\eq{
		&\hspace{-8mm}\Ex\|\tw_0^{t+1}\|^2 + \gamma(a_{t+1}^2+b_{t}^2)\nn\\
		\leq &\;
		\left(1-\frac{3}{4}\mu\nu N\right) \Ex\|\tw_0^t\|^2 + 4\mu N \frac{\delta^2}{\nu}(a_t^2+b_{t-1}^2)\nn\\
		&\;{} + 5\gamma\delta^2\mu^2N^2(a_{t+1}^2+b_{t}^2) + 3\gamma\delta^2 \mu^2 N^2\Ex\|\tw_0^{t+1}\|^2\nn\\
		&\;{}+5\gamma\delta^2\mu^2N^2(a_t^2+b_{t-1}^2) + 3\gamma\delta^2 \mu^2 N^2\Ex\|\tw_0^t\|^2 \label{vnuisd3}
	}
	which simplifies to
	\eq{
		&\hspace{-3mm}(1-3\gamma\delta^2 \mu^2N^2)\Ex\|\tw_0^{t+1}\|^2 + \gamma(1-5\delta^2\mu^2N^2)(a_{t+1}^2+b_{t}^2) \nn\\
		&\leq
		\left(1-\frac{3}{4}\mu\nu N + 3\gamma\delta^2 \mu^2 N^2\right)\Ex\|\tw_0^t\|^2
		\nn\\
		&\hspace{5mm}+\left(4\mu N \frac{\delta^2}{\nu} + 5\gamma\delta^2\mu^2N^2\right) (a_t^2+b_{t-1}^2)
	}
	Under the condition $1-3\gamma\delta^2 \mu^2N^2 > 0$, which is equivalent to
	\eq{
		\boxed{{\rm Condition\ \#2:}\;\; \mu^2\gamma < \frac{1}{3\delta^2  N^2}}
	}
	it holds that
	\eq{
		&\hspace{-3mm}\Ex\|\tw_0^{t+1}\|^2 + \gamma\frac{1-5\delta^2\mu^2N^2}{1-3\gamma\delta^2 \mu^2N^2}(a_{t+1}^2+b_{t}^2) \nn\\
		&\leq
		\frac{1-3\mu\nu N / 4 + 3\gamma\delta^2 \mu^2 N^2}{1-3\gamma\delta^2 \mu^2N^2}\Ex\|\tw_0^t\|^2
		\nn\\
		&\hspace{5mm}+\frac{4\mu N \frac{\delta^2}{\nu} +5\gamma\delta^2\mu^2N^2}{1-3\gamma\delta^2 \mu^2N^2} (a_t^2+b_{t-1}^2)\nn\\
		&=
		\frac{1-3\mu\nu N / 4 + 3\gamma\delta^2 \mu^2 N^2}{1-3\gamma\delta^2 \mu^2N^2}\times
		\nn\\
		&\hspace{6mm}\left(\Ex\|\tw_0^t\|^2+\frac{4\mu N \frac{\delta^2}{\nu} +5\gamma\delta^2\mu^2N^2}{1-3\mu\nu N/4 + 3\gamma\delta^2 \mu^2 N^2} (a_t^2+b_{t-1}^2)\right)
	}
	
	This relation in turn implies that
	\eq{
		&\hspace{-1mm}\Ex\|\tw_0^{t+1}\|^2 + \gamma(1-5\delta^2\mu^2N^2)(a^2_{t+1}+b^2_t) \nn\\
		&\leq
		\frac{1-3\mu\nu N / 4 + 3\gamma\delta^2 \mu^2 N^2}{1-3\gamma\delta^2 \mu^2N^2}\times
		\nn\\
		&\hspace{6mm}\left(\Ex\|\tw_0^t\|^2
		+\frac{4\mu N \frac{\delta^2}{\nu} +5\gamma\delta^2\mu^2N^2}{1-3\mu\nu N/4 + 3\gamma\delta^2 \mu^2 N^2} (a_t^2+b_{t-1}^2)\right)\label{2389hg32.3}
	}
	
	We can again simplify the result by noting that
	\eq{
		1-3\mu\nu N / 4 + 3\gamma\delta^2 \mu^2 N^2 =& 1-\mu\nu N / 4 - ( \mu\nu N / 2  - 3\gamma\delta^2 \mu^2 N^2)\nn\\
		\le& 1-\mu\nu N / 4 \label{38g.1}
	}
	where the inequality holds when  $\mu\nu N / 2  - 3\gamma\delta^2 \mu^2 N^2 \ge 0$, i.e.,
	\eq{
		\boxed{{\rm Condition\ \#3:}\;\; \mu\gamma \le \frac{\nu}{6 \delta^2 N}}
		\label{189g31.23}
	}
	In addition, we have the lower bound
	\eq{
		1-\frac{3}{4}\mu\nu N + 3\gamma\delta^2 \mu^2 N^2 \ge 1-\frac{3}{4}\mu\nu N
	}
	Using condition \#1 from Eq. (\ref{23ns3299}), we have
	\eq{
		1-\frac{3}{4}\mu\nu N \geq 1 - \frac{3\nu^2}{16\delta^2}\geq \frac{13}{16} \label{38g.2}
	}
	In a similar manner,
	\eq{
		4\mu N \frac{\delta^2}{\nu} +5\gamma\delta^2\mu^2N^2 \le&\; 4\mu N \frac{\delta^2}{\nu} + \mu N \frac{\delta^2}{\nu}\nn\\
		=&\; 5\mu N \frac{\delta^2}{\nu}\label{38g.3}
	}
	where the last inequality holds when $ \mu\gamma \le \frac{1}{5 \nu N} $, which is always valid under condition \#3 in Eq. (\ref{189g31.23}) since the latter implies that $\mu\gamma \le \frac{1}{6\delta N}$. Substituting (\ref{38g.1}), (\ref{38g.2}), and (\ref{38g.3}) into (\ref{2389hg32.3}), we find that
	\eq{
		&\hspace{-0mm}\Ex\|\tw_0^{t+1}\|^2 + \gamma(1-5\delta^2\mu^2N^2)(a_{t+1}^2+b_{t}^2) \nn\\
		&\;\leq
		\frac{1-\mu\nu N / 4}{1-3\gamma\delta^2 \mu^2N^2}\left(\Ex\|\tw_0^t\|^2
		+ \frac{16}{13}\cdot 5\mu N \frac{\delta^2}{\nu} (a_t^2+b_{t-1}^2)\right)
	}
	Under condition \#1 in Eq. (\ref{23ns3299}), we have
	\eq{
		1-5\delta^2\mu^2N^2 \ge 1-5\delta^2N^2\frac{\nu^2}{16\delta^4N^2}\geq 1-\frac{5}{16}\geq \frac{11}{16}
	}
	and, hence,
	\eq{
		&\hspace{-5mm}\Ex\|\tw_0^{t+1}\|^2 + \frac{11}{16}\gamma(a_{t+1}^2+b_{t}^2)\nnb
		\leq&\;
		\frac{1-\mu\nu N / 4}{1-3\gamma\delta^2 \mu^2N^2}\left(\Ex\|\tw_0^t\|^2
		+ \frac{80}{13} \mu N \frac{\delta^2}{\nu} (a_t^2+b_{t-1}^2)\right)\nn\\
		\leq&\;
		\frac{1-\mu\nu N / 4}{1-3\gamma\delta^2 \mu^2N^2}\left(\Ex\|\tw_0^t\|^2
		+ \frac{99}{16} \mu N \frac{\delta^2}{\nu} (a_t^2+b_{t-1}^2)\right)
	}
	where the last inequality is unnecessary but is introduced for convenience. Recall that we are free to choose $\gamma$, so assume we choose it to satisfy
	\eq{
		\frac{11}{16} \gamma = \frac{99}{16} \mu N \frac{\delta^2}{\nu}\;\Longrightarrow \;
		\gamma = 9\mu N \frac{\delta^2}{\nu}
		\label{f289g.23}
	}
	It then follows that:
	\eq{
		&\hspace{-2mm}\Ex\|\tw_0^{t+1}\|^2 + \frac{11}{16}\gamma(a_{t+1}^2+b_{t}^2)\nn\\
		&\leq
		\frac{1-\mu\nu N / 4}{1-3\gamma\delta^2 \mu^2N^2}\left(\Ex\|\tw_0^t\|^2
		+ \frac{11}{16} \gamma (a_t^2+b_{t-1}^2)\right)\nn\\
		&\define
		\alpha \left(\Ex\|\tw_0^t\|^2
		+ \frac{11}{16} \gamma (a_t^2+b_{t-1}^2)\right)
	}
	where we introduced the positive parameter
	\eq{
		\alpha \define \frac{1-\mu\nu N / 4}{1-3\gamma\delta^2 \mu^2N^2}\label{alpha.eq}
	}
	This parameter controls the speed of convergence. It will hold that $\alpha<1$ when
	\eq{
		\frac{1-\mu\nu N / 4}{1-3\gamma\delta^2 \mu^2N^2} = \frac{1-\mu\nu N / 4}{1-27\delta^4 \mu^3 N^3 / \nu} < 1
		\;\Longleftrightarrow \; \mu < \sqrt{\frac{1}{108}}\frac{\nu}{\delta^2 N}\label{23nbs9}
	}
	Let us now re-examine conditions \#1 through \#3, along with (\ref{23nbs9}), when $\gamma$ is chosen according to
	(\ref{f289g.23}). In this case, conditions \#1 through \#3 become
	\eq{
		{\rm Conditions\ \#1\ to\ \#3}:\mu \!\leq\! \frac{\nu}{4\delta^2 N},\;
		\mu^3\!<\! \frac{\nu}{27 \delta^4  N^3},\;\;
		\mu^2\!\le\! \frac{\nu^2}{54 \delta^4 N^2}
	}
	which can be met by:
	\eq{
		\boxed{
			\mu \leq \frac{\nu}{4\delta^2 N},\;\;
			\mu < \frac{1}{3\delta N}\left(\frac{\nu}{\delta}\right)^{1/3},\;\;
			\mu \le \sqrt{\frac{1}{54}}\frac{\nu}{\delta^2 N}}\label{2389hg2}
	}
	All three conditions and condition \eqref{23nbs9} can be satisfied by the following single sufficient bound on the step-size parameter (since $11^2>108$):
	\eq{
		\boxed{\mu \leq \frac{\nu}{11\delta^2 N}}
		\label{189jf2.3}
	}
	
	\section{Proof of lemma \ref{lemma.avrg}}\label{app.lemma.avrg}
	Subtracting $w^\star$ from both sides of (\ref{eq.avrg.main}), we obtain:
	\eq{
		\tw^{t+1}_0
		=&\tw^t_0 + \mu  N \nabla J(\w_0^t)+\mu\! \sum_{i=0}^{N-1}\! \left[\nabla Q(\w_{i}^t;x_{\n})- \nabla Q(\w^{t}_0;x_{\n}) \right]\nn\\
		&\;\;-\mu \sum_{i=0}^{N-1}\left[\nabla Q(\w^{t}_0;x_{\n'})-\nabla Q(\w^{t-1}_i;x_{\n'})\right]
	}
	Then, taking the squared norm and applying Jensen's inequality, we establish the first recursion for any $t\in (0,1)$:
	\eq{
		\|\tw^{t+1}_0\|^2 \leq\;& \frac{1}{t}\left\|\tw^{t}_0+\mu N\nabla J(\w^t_0)\right\|^2\nn\\
		&\;\;+\frac{2\mu^2}{1-t} \Big\|\sum_{i=0}^{N-1} \left[\nabla Q(\w_{i}^t;x_{\n})-\nabla Q(\w^{t}_0;x_\n)\right]\Big\|^2\nn\\
		&\;\;{}+\frac{2\mu^2}{1-t} \Big\|\sum_{i=0}^{N-1} \left[\nabla Q(\w_{0}^t;x_{\n'})-\nabla Q(\w^{t-1}_i;x_{\n'})\right]\Big\|^2\nn\\
		\leq\;&\frac{1}{t}\|\tw^{t}_0+\mu N\nabla J(\w^t_0)\|^2
		+\frac{2\mu^2\delta^2 N}{1-t} \sum_{i=0}^{N-1}\| \w_{i}^t-\w^{t}_0\|^2\nn\\
		&
		+\frac{2\mu^2\delta^2 N}{1-t} \sum_{i=0}^{N-1}\| \w_{N}^{t-1}-\w^{t-1}_{i}\|^2
	}
	Using an argument similar to (\ref{f9j30.3}) and letting $t=1-\mu N\nu$,  assuming $\mu\leq 1/(N\nu)$, we obtain:
	\eq{
		\|\tw^{t+1}_0\|^2 &\leq\left(1-\frac{\mu\nu N - \mu^2N^2\delta^2}{1-\mu N\nu}\right)\|\tw^{t}_0\|^2\nn\\
		&\hspace{5mm} + \frac{2\mu\delta^2 }{\nu} \left( \sum_{i=0}^{N-1}\| \w_{i}^t-\w^{t}_0\|^2+\sum_{i=0}^{N-1}\| \w_{N}^{t-1}-\w^{t-1}_{i}\|^2\right)
	}
	Taking the expectation of both sides, we establish \eqref{38jgg3g}. The forward inner difference recursion can be obtain by following the same procedure as in \eqref{289jg3}:
	\eq{
		&\hspace{-1mm}\|\w^t_i - \w^t_0\|^2 \nn\\
		\leq &\;i \sum_{m=0}^{i-1} \|\w^t_{m+1} - \w^t_m\|^2\nn\\
		\stackrel{\eqref{2389hg.23g}}{\leq}&\;3\mu^2\delta^2i\sum_{m=0}^{i-1}\!\left(\!\|\w^t_m-\w^t_0\|^2 \!+\! \frac{1}{N}\sum_{n'=0}^{N-1}\|\w^{t-1}_{n'}\!-\!\w^{t-1}_N\|^2\!+\!\|\tw^{t}_0\|^2\!\right)\nn\\
		=&\;3\mu^2\delta^2i\sum_{m=0}^{i-1}\|\w^t_m-\w^t_0\|^2\nn\\
		&\;\;\;\;{} + 3\mu^2\delta^2i^2\left(\frac{1}{N}\sum_{n'=0}^{N-1}\|\w^{t-1}_{n'}-\w^{t-1}_N\|^2 + \|\tw_0^t\|^2\right)
	}
	\noindent Summing over $i$, we have
	\eq{
		&\hspace{-4mm}\sum_{i=0}^{N-1}\|\w_i^t - \w_0^t\|^2\nn\\
		\leq&\; 3\mu^2\delta^2 \left( \sum_{i=0}^{N-1}i\sum_{m=0}^{i-1}\|\w_m^t - \w_0^t\|^2 \right.\nn\\
		&\hspace{12mm}\left.+ \sum_{i=0}^{N-1}i^2\left(\frac{1}{N}\sum_{n'=0}^{N-1}\|\w^{t-1}_{n'}-\w^{t-1}_N\|^2 + \|\tw_0^t\|^2\right) \right) \nnb
		=&\; 3\mu^2\delta^2\left(\sum_{m=0}^{N-1}\sum_{i=m+1}^{N-1}i\|\w_m^t - \w_0^t\|^2 \right.\nn\\
		&\hspace{12mm}\left.+\sum_{i=0}^{N-1}i^2 \left(\frac{1}{N}\sum_{n'=0}^{N-1}\|\w^{t-1}_{n'}-\w^{t-1}_N\|^2 + \|\tw_0^t\|^2\right)\right) \nnb
		\stackrel{(a)}{\leq}&\; 3\mu^2\delta^2N^2 \sum_{m=0}^{N-1}\|\w_{m}^t-\w_0^t \|^2\nn\\
		&\;\;\; + \mu^2 \delta^2 N^2\left(\sum_{n'=0}^{N-1}\|\w_N^{t-1}- \w_{n'}^{t-1}\|^2 + N\|\tw_0^t\|^2\right)\nnb
		=&\; 3\mu^2\delta^2N^2 \sum_{i=0}^{N-1}\|\w_{i}^t-\w_0^t \|^2\nn\\
		&\;\;\; + \mu^2 \delta^2 N^2\left(\sum_{i=0}^{N-1}\|\w_N^{t-1}- \w_{i}^{t-1}\|^2 + N\|\tw_0^t\|^2\right)
	}
	where step (a) is because:
	\eq{
		\sum_{m+1}^{N-1} i \leq N^2,\; \;\;\sum_{i=0}^{N-1} i^2 = \frac{(N-1)N(2N-1)}{6}\leq \frac{N^3}{3}
	}
	
	Lastly, we establish the backwards inner difference term using the same argument as in (\ref{13jgds.ge}):
	\eq{
		&\hspace{-3mm}\| \w_{N}^{t}-\w^{t}_{i}\|^2 \nn\\
		=&\; \|\w_{N}^{t} - \w_{N-1}^{t} + \w_{N-1}^{t} - \cdots + \w^{t}_{i+1} - \w^t_i\|^2 \nn\\
		\le &\ (N-i)\sum_{m=i}^{N-1}\|\w_{m+1}^t - \w_{m}^t\|^2\nn\\
		\le &\ 3\mu^2\delta^2(N-i)\sum_{m=i}^{N-1}\Bigg(\|\w^t_m-\w^t_0\|^2 \nn\\
		&\hspace{27mm}\left.+ \frac{1}{N}\sum_{n'=0}^{N-1}\|\w^{t-1}_{n'}-\w^{t-1}_N\|^2+\|\tw^{t}_0\|^2\right) \nnb
		\le &\ 3\mu^2\delta^2(N-i)\sum_{m=i}^{N-1}\|\w^t_m-\w^t_0\|^2 \nn\\
		&\;+ \frac{3\mu^2\delta^2(N-i)^2}{N}\sum_{n'=0}^{N-1}\|\w^{t-1}_{n'}-\w^{t-1}_N\|^2 \!+\! 3\mu^2\delta^2(N-i)^2 \|\tw^{t}_0\|^2
	}
	Observing that this backward term is summing from $0$ to $N-1$, rather than from $1$ to $N-1$ as in SAGA with RR, we have
	\eq{
		&\hspace{-3mm} \sum_{i=0}^{N-1}\| \w_{N}^{t}-\w^{t}_{i}\|^2 \nnb
		\le &\ 3\mu^2\delta^2\sum_{i=0}^{N-1}(N-i)\sum_{m=i}^{N-1}\|\w^t_m-\w^t_0\|^2 \nn\\
		&\;\;{} +3\mu^2\delta^2\sum_{i=0}^{N-1}(N-i)^2\!\left(\!\frac{1}{N}\sum_{n'=0}^{N-1}\|\w^{t-1}_{n'}-\w^{t-1}_N\|^2+\|\tw^{t}_0\|^2\! \right) \nnb
		=&\ 3\mu^2\delta^2\sum_{m=0}^{N-1}\sum_{i=0}^{m}(N-i)\|\w^t_m-\w^t_0\|^2 \nn\\
		&\;\;{}+ 3\mu^2\delta^2\frac{N(N\!+\!1)(2N\!+\!1)}{6}\!\left(\!\frac{1}{N}\!\sum_{n'=0}^{N-1}\|\w^{t-1}_{n'}\!-\!\w^{t-1}_N\|^2\!+\!\|\tw^{t}_0\|^2 \!\right) \nnb
		\le&\ 3\mu^2\delta^2N^2\sum_{m=0}^{N-1}\|\w^t_m-\w^t_0\|^2 \nn\\
		&\;\;{}+ 3\mu^2\delta^2N^2 \left( \sum_{n'=0}^{N-1}\|\w^{t-1}_{n'}-\w^{t-1}_N\|^2 + N \|\tw_0^t\|^2 \right)\nn\\
		=&\ 3\mu^2\delta^2N^2\sum_{i=0}^{N-1}\|\w^t_i-\w^t_0\|^2\nn\\
		&\;\;{}+ 3\mu^2\delta^2N^2 \left( \sum_{i=0}^{N-1}\|\w^{t-1}_i-\w^{t-1}_N\|^2 + N \|\tw_0^t\|^2 \right)
	}
	where in the last inequality we used the fact that
	\eq{
		\frac{N(N+1)(2N+1)}{6} \le N^3, \;\;\;\;\forall N
	}
	
	\section{Proof of theorem \ref{theorem.2}}\label{app.theorem.2}
	We let
	\eq{
		a_t \define \frac{1}{N}\sum_{i=0}^{N-1}\Ex \| \w_{i}^t-\w^{t}_0\|^2,\;\;\; b_t \define \frac{1}{N}\sum_{i=0}^{N-1} \Ex\| \w_{N}^t-\w^{t}_i\|^2
	}
	The recursions available so far for AVRG are:
	\eq{
		\Ex\|\tw^{t+1}\|^2\leq&\left(1-\frac{\mu\nu N - \mu^2N^2\delta^2}{1-\mu N\nu}\right)\Ex\|\tw^{t}_0\|^2 + \frac{2\mu\delta^2N}{\nu} \left( a_t +b_{t-1}\right)
		\\
		a_{t+1}\leq&\; 3\mu^2\delta^2N^2a_{t+1} + \mu^2 \delta^2 N^2b_t+  \mu^2 \delta^2 N^2\Ex\|\tw_0^{t+1}\|^2\label{23gj23g}
		\\
		b_t \leq&\;  3\mu^2\delta^2N^2(a_t+b_{t-1}) + 3\mu^2\delta^2N^2 \Ex\|\tw_0^t\|^2
	}
	
	which have exactly the same form as recursions \eqref{z23n99}---\eqref{1389jg2.1} except for the coefficients. To simplify the argument, we can replace (\ref{23gj23g}) by:
	\eq{
		a_{t+1}\leq&\; 3\mu^2\delta^2N^2(a_{t+1} +b_t) +  \mu^2 \delta^2 N^2\Ex\|\tw_0^{t+1}\|^2
	}
	Similar to the derivation of \eqref{vnuisd3}, we  have:
	
	\eq{
		&\hspace{-1cm}(1-\gamma\mu^2\delta^2N^2)\Ex\|\tw^{t+1}\|^2 + \gamma(1-3\mu^3\delta^2N^2)(a_{t+1}+b_t)\nn\\
		\leq&\, \left(1-\frac{3\mu\nu N}{4}\right)\Ex\|\tw^{t}_0\|^2 + \frac{2\mu\delta^2N}{\nu} \left( a_t +b_{t-1}\right)\nn\\
		&\;\;+\gamma \Big(3\mu^2\delta^2N^2(a_t+b_{t-1}) + 3\mu^2\delta^2N^2 \Ex\|\tw_0^t\|^2 \Big)\nn\\
		=&\, \left(1-\frac{3\mu\nu N}{4}+3\gamma\mu^2\delta^2N^2\right)\Ex\|\tw^{t}_0\|^2 \nn\\
		&\;\;\;\;+ \left(\frac{2\mu\delta^2N}{\nu}+3\gamma\mu^2\delta^2N^2\right) \left( a_t +b_{t-1}\right)
	}
	under
	\eq{
		\boxed{{\rm Condition\ \#1:}\;\; \mu \leq \frac{\nu}{4\delta^2N}}
	}
	Under the condition $1-\gamma\delta^2 \mu^2N^2 > 0$, which is equivalent to
	\eq{
		\boxed{{\rm Condition\ \#2:}\;\; \mu^2\gamma < \frac{1}{\delta^2  N^2}}
	}
	it further holds that
	\eq{
		&\hspace{-4mm}\Ex\|\tw^{t+1}\|^2 + \gamma(1-3\mu^2\delta^2N^2)(a_{t+1}+b_t)\nn\\
		\leq&\, \frac{1-3\mu\nu N/4+3\gamma\mu^2\delta^2N^2}{1-\gamma\delta^2\mu^2N^2}\times\nn\\
		&\;\left(\Ex\|\tw^{t}_0\|^2 + \frac{\frac{2\mu\delta^2N}{\nu}+3\gamma\mu^2\delta^2N^2}{1-3\mu\nu N/4+3\gamma\mu^2\delta^2N^2} \left( a_t +b_{t-1}\right)\right)
	}
	Note that the numerator $1-3\mu\nu N/4+3\gamma\mu^2\delta^2N^2$ is the same as SAGA in (\ref{38g.1}). Thus, under condition:
	\eq{
		\boxed{{\rm Condition\ \#3:}\;\; \mu\gamma \le \frac{\nu}{6 \delta^2 N}}
	}
	we have:
	\eq{
		\frac{11}{16}\leq\;1-3\mu\nu N / 4 + 3\gamma\delta^2 \mu^2 N^2
		\le&\; 1-\mu\nu N / 4
	}
	Lastly, we can verify that
	\eq{
		\frac{2\mu\delta^2N}{\nu}+3\gamma\mu^2\delta^2N^2\leq &\; 
		\frac{2\mu\delta^2N}{\nu} + \frac{\mu\delta^2N}{\nu}\nn\\
		\leq&\; \frac{3\mu\delta^2N}{\nu}
	}
	where the last inequality holds when $\mu\gamma \leq \frac{1}{3\nu N}$, which is always valid under condition \#3. Now, collecting the results, we have
	\eq{
		&\hspace{-4mm}\Ex\|\tw^{t+1}\|^2 + \gamma\frac{13}{16}(a_{t+1}+b_t)\nn\\
		\leq&\, \frac{1-\mu\nu N/4}{1-\gamma\mu^2\delta^2N^2}\left(\Ex\|\tw^{t}_0\|^2 + 3\frac{16}{11}\mu N\frac{\delta^2}{\nu} \left( a_t +b_{t-1}\right)\right)\nn\\
		\leq&\, \frac{1-\mu\nu N/4}{1-\gamma\mu^2\delta^2N^2}\left(\Ex\|\tw^{t}_0\|^2 + 3\frac{26}{16}\mu N\frac{\delta^2}{\nu} \left( a_t +b_{t-1}\right)\right)
	}
	Assume we choose $\gamma$ such that
	\eq{
		\gamma\frac{13}{16} = 3\frac{26}{16}\mu N \frac{\delta^2}{\nu} \;\;\Longrightarrow\;\; \gamma = 6 \mu N \frac{\delta^2}{\nu}\label{f289g.232}
	}
	It then follows that:
	\eq{
		&\hspace{-3mm}\Ex\|\tw_0^{t+1}\|^2 + \frac{13}{16}\gamma(a_{t+1}^2+b_{t}^2)\nn\\
		&\leq
		\frac{1-\mu\nu N / 4}{1-3\gamma\delta^2 \mu^2N^2}\left(\Ex\|\tw_0^t\|^2
		+ \frac{13}{16} \gamma (a_t^2+b_{t-1}^2)\right)\nn\\
		&\define
		\alpha \left(\Ex\|\tw_0^t\|^2
		+ \frac{13}{16} \gamma (a_t^2+b_{t-1}^2)\right)
	}
	where we introduced the positive parameter
	\eq{
		\alpha \define \frac{1-\mu\nu N / 4}{1-3\gamma\delta^2 \mu^2N^2}\label{alpha.eq2}
	}
	This parameter satisfies $\alpha<1$ for
	\eq{
		\frac{1-\mu\nu N / 4}{1-3\gamma\delta^2 \mu^2N^2} = \frac{1-\mu\nu N / 4}{1-18\delta^4 \mu^3 N^3/ \nu} < 1
		\;\Longleftrightarrow \; \mu < \sqrt{\frac{1}{72}}\frac{\nu}{\delta^2 N}\label{23nbs9.2}
	}
	
	We re-examine conditions \#1--\#3 when $\gamma$ is chosen according to
	(\ref{f289g.232}). In this case, these conditions become
	\eq{
		{\rm Conditions\ \#1\ to\  \#3}:\mu \!\leq\! \frac{\nu}{4\delta^2 N},\;
		\mu^3\!<\! \frac{\nu}{6 \delta^4  N^3},\;\;
		\mu^2\!\le\! \frac{\nu^2}{36\delta^4 N^2}
	}
	which can be met by:
	\eq{
		\boxed{
			\mu \leq \frac{\nu}{4\delta^2 N},\;\;
			\mu < \frac{1}{2\delta N}\left(\frac{\nu}{\delta}\right)^{1/3},\;\;
			\mu \le \frac{\nu}{6\delta^2 N}}
	}
	
	All these three conditions and the condition for $\alpha<1$ can be satisfied by the following single sufficient bound on the step-size parameter:
	\eq{
		\boxed{\mu \leq \frac{\nu}{9\delta^2 N}}
	}
\normalsize
\bibliographystyle{IEEEbib}
\bibliography{saga_rr}

\begin{thebibliography}{10}

\bibitem{johnson2013accelerating}
R.~Johnson and T.~Zhang,
\newblock ``Accelerating stochastic gradient descent using predictive variance
  reduction,''
\newblock in {\em Proc. Advances in Neural Information Processing Systems {\rm
  (NIPS)}}, Lake Tahoe, Navada, 2013, pp. 315--323.

\bibitem{defazio2014saga}
A.~Defazio, F.~Bach, and S.~Lacoste-Julien,
\newblock ``{SAGA}: A fast incremental gradient method with support for
  non-strongly convex composite objectives,''
\newblock in {\em Proc. Advances in Neural Information Processing Systems {\rm
  (NIPS)}}, Montreal, Canada, 2014, pp. 1646--1654.

\bibitem{defazio2014finito}
A.~Defazio, J.~Domke, and T.~S. Caetano,
\newblock ``Finito: A faster, permutable incremental gradient method for big
  data problems.,''
\newblock in {\em Proc. International Conference of Machine Learning {\rm
  (ICML)}}, Beijing, China, 2014, pp. 1125--1133.

\bibitem{shalev2013stochastic}
S.~Shalev-Shwartz and Tong Zhang,
\newblock ``Stochastic dual coordinate ascent methods for regularized loss,''
\newblock {\em Journal of Machine Learning Research}, vol. 14, no. 1, pp.
  567--599, 2013.

\bibitem{roux2012stochastic}
N.~L. Roux, M.~Schmidt, and F.~R. Bach,
\newblock ``A stochastic gradient method with an exponential convergence rate
  for finite training sets,''
\newblock in {\em Proc. Advances in Neural Information Processing Systems {\rm
  (NIPS)}}, Lake Tahoe, Navada, 2012, pp. 2663--2671.

\bibitem{bottou2009curiously}
L.~Bottou,
\newblock ``Curiously fast convergence of some stochastic gradient descent
  algorithms,''
\newblock in {\em Proc. Symposium on Learning and Data Science}, Paris, 2009,
  pp. 1--5.

\bibitem{recht2012toward}
B.~Recht and C.~R{\'e},
\newblock ``Toward a noncommutative arithmetic-geometric mean inequality:
  Conjectures, case-studies, and consequences,''
\newblock in {\em Proc. Conference on Learning Theory {\rm (COLT)}}, Edinburgh,
  Scotland, 2012, pp. 1--11.

\bibitem{gurbuzbalaban2015random}
M.~G{\"u}rb{\"u}zbalaban, A.~Ozdaglar, and P.~Parrilo,
\newblock ``Why random reshuffling beats stochastic gradient descent,''
\newblock {\em arXiv:1510.08560}, Oct. 2015.

\bibitem{nesterov2013introductory}
Y.~Nesterov,
\newblock {\em Introductory Lectures on Convex Optimization: A basic course},
  vol.~87,
\newblock Springer, 2013.

\bibitem{polyak1987introduction}
B.~T. Polyak,
\newblock {\em Introduction to Optimization},
\newblock Optimization Software, NY, 1987.

\bibitem{ying2017rr}
B.~Ying, K.~Yuan, S.~Vlaski, and A.~H. Sayed,
\newblock ``On the performance of random reshuffling in stochastic learning,''
\newblock in {\em Proc. Information Theory and Applications Workshop {\rm
  (ITA)}}, San Diego, CA, Feb. 2017, pp. 1--5.

\bibitem{sayed2014adaptation}
A.~H. Sayed,
\newblock ``Adaptation, learning, and optimization over networks,''
\newblock {\em Foundations and Trends in Machine Learning}, vol. 7, no. 4--5,
  pp. 311--801, 2014.

\bibitem{de2016efficient}
S.~De and T.~Goldstein,
\newblock ``Efficient distributed {SGD} with variance reduction,''
\newblock in {\em Proc. IEEE International Conference on Data Mining (ICDM)},
  Barcelona, Spain, 2016, pp. 111--120.

\bibitem{gurbuzbalaban2015convergence}
M.~G{\"u}rb{\"u}zbalaban, A.~Ozdaglar, and P.~Parrilo,
\newblock ``Convergence rate of incremental gradient and newton methods,''
\newblock {\em arXiv:1510.08562}, Oct. 2015.

\bibitem{Shamir2016Without}
O.~Shamir,
\newblock ``Without-replacement sampling for stochastic gradient methods:
  Convergence results and application to distributed optimization,''
\newblock {\em arXiv:1603.00570}, Mar. 2016.

\bibitem{NIPS2015_5821}
S.~J.~Reddi, A.~Hefny, S.~Sra, B.~Poczos, and A.~J. Smola,
\newblock ``On variance reduction in stochastic gradient descent and its
  asynchronous variants,''
\newblock in {\em Advances in Neural Information Processing {\rm (NIPS)}}, pp.
  2647--2655. Montréal, Canada, 2015.

\bibitem{konevcny2016federated}
J.~Konecny, H.B. McMahan, D.~Ramage, and P.~Richt{\'a}rik,
\newblock ``Federated optimization: distributed machine learning for on-device
  intelligence,''
\newblock {\em available on arXiv:1610.02527}, Oct. 2016.

\bibitem{yuan2017efficient}
K.~Yuan, B.~Ying, and A.~H. Sayed,
\newblock ``Efficient variance-reduced learning for fully decentralized
  on-device intelligence,''
\newblock {\em available on arXiv:1708.01384}, August 2017.

\bibitem{NIPS2015_5711}
R.~Harikandeh, M.~O. Ahmed, A.~Virani, M.~Schmidt, J.~Konecny, and S.~Sallinen,
\newblock ``Stop wasting my gradients: Practical {SVRG},''
\newblock in {\em Advances in Neural Information Processing {\rm (NIPS)}}, pp.
  2251--2259. Montréal, Canada, 2015.

\bibitem{xiao2014proximal}
L.~Xiao and T.~Zhang,
\newblock ``A proximal stochastic gradient method with progressive variance
  reduction,''
\newblock {\em SIAM Journal on Optimization}, vol. 24, no. 4, pp. 2057--2075,
  2014.

\end{thebibliography}

\end{document}